%% file: main_tmlr.tex
\documentclass[10pt]{article} 
\usepackage[preprint]{tmlr_style/tmlr}

\definecolor{recourseColor}{HTML}{2563EB}
\definecolor{benchColor}{HTML}{7C3AED}
\newcommand{\RecourseBench}{%
\textbf{\textcolor{recourseColor}{Recourse}\kern0.08em
\textcolor{benchColor}{Bench}}\xspace}

\newcommand{\numMethodsFull}{27\xspace}

\input{feedback_tools}
\usepackage{needspace}

\usepackage[utf8]{inputenc} 
\usepackage[T1]{fontenc}    
\usepackage{hyperref}       
\usepackage{url}            
\usepackage{booktabs}       
\usepackage{amsfonts}       
\usepackage{nicefrac}       
\usepackage{microtype}      
\usepackage{xcolor}         
\usepackage{amsmath}
\usepackage{cleveref} 
\usepackage{graphicx}
\usepackage{multirow}
\usepackage[table]{xcolor}
\usepackage{longtable}
\usepackage{caption}
\usepackage{float}

\usepackage{tikz}
\usepackage{enumitem}
\setlist{
  topsep=1pt,       
  partopsep=0pt,    
  itemsep=1.5pt,      
  parsep=0pt        
}
\usetikzlibrary{matrix, positioning}
\usepackage{subcaption}
\usepackage{xspace}
\input{tmlr_style/math_commands.tex}

\usepackage{hyperref}
\usepackage{url}

\title{\RecourseBench: A Modular Framework for Reproducible \\ Algorithmic Recourse Evaluation}

\author{\name Hashir Ahmed \email  h239ahme@uwaterloo.ca \\
      \addr Department of Electrical and Computer Engineering\\
      University of Waterloo
      \AND
      \name Zahra Khotanlou \email  zkhotanlou@uwaterloo.ca \\
      \addr Department of Electrical and Computer Engineering\\
      University of Waterloo
      \AND
      \name Chenghao Tan \email  c27tan@uwaterloo.ca \\
      \addr Department of Electrical and Computer Engineering\\
      University of Waterloo
      \AND
      \name Ahmed Abdelaal \email  a49abdel@uwaterloo.ca \\
      \addr Department of Electrical and Computer Engineering\\
      University of Waterloo
      \AND
      \name Amir-Hossein Karimi \email  amirh.karimi@uwaterloo.ca \\
      \addr Department of Electrical and Computer Engineering\\
      University of Waterloo
      }

\def\month{MM}  
\def\year{YYYY} 
\def\openreview{\url{https://openreview.net/forum?id=XXXX}} 

\begin{document}

\maketitle

\begin{abstract}

\input{sections/Abstract}
\end{abstract}

\input{sections/Introduction}
\input{sections/Preliminaries}

\input{sections/Framework}

\input{sections/Reproduction}

\input{sections/Evaluation}

\input{sections/Discussion}
\input{sections/Conclusion}


\bibliography{references, zahra_references}
\bibliographystyle{plainnat}

\clearpage

\appendix
\input{sections/Appendix}

\end{document}

%% file: feedback_tools.tex
\usepackage{xcolor}
\usepackage{soul}
\usepackage{tikz}
\usepackage{etoolbox}
\usetikzlibrary{calc, arrows.meta, backgrounds}

\pgfdeclarelayer{amirbackground}
\pgfsetlayers{amirbackground,main}

\definecolor{amirHighlight}{HTML}{F9A8D4} 
\definecolor{amirBox}{HTML}{FDF2F8}       
\definecolor{amirBorder}{HTML}{DB2777}    
\definecolor{amirText}{HTML}{3B1025}      
\definecolor{amirLine}{HTML}{EC4899}      

\sethlcolor{amirHighlight}

\newlength{\amirNoteWidth}
\usepackage[colorinlistoftodos, obeyFinal]{todonotes}
\tikzset{
  amirConnector/.style={
    draw=amirLine,
    line width=0.8pt,
    dash pattern=on 2pt off 3pt,
    opacity=0.45,
    line cap=round,
    line join=round
  }
}

\newcounter{amirNoteSide}

\newcount\amirDesiredSide
\makeatletter

\newdimen\amir@leftMparBottom
\newdimen\amir@rightMparBottom

\patchcmd{\@addmarginpar}
{\ifnum \@tempcnta <\z@  \global\setbox\@marbox\box\@currbox \fi}
{%
  \ifnum\count\@marbox=-2\relax
    \@tempcnta\m@ne
  \else\ifnum\count\@marbox=-3\relax
    \@tempcnta\@ne
  \fi\fi
  \ifnum \@tempcnta <\z@
    \global\setbox\@marbox\box\@currbox
  \fi
}
{}
{%
  \PackageError{feedback_tools}
    {Could not patch margin-note side selection}
    {The LaTeX marginpar implementation may have changed.}%
}

\patchcmd{\@addmarginpar}
{\@tempdima\@mparbottom}
{%
  \ifnum\@tempcnta<\z@
    \let\amir@activeMparBottom\amir@leftMparBottom
  \else
    \let\amir@activeMparBottom\amir@rightMparBottom
  \fi
  \@tempdima\amir@activeMparBottom
}
{}
{%
  \PackageError{feedback_tools}
    {Could not patch margin-note stack selection}
    {The LaTeX marginpar implementation may have changed.}%
}

\patchcmd{\@addmarginpar}
{%
    \global\@mparbottom\@pageht
    \global\advance\@mparbottom\@tempdima
    \global\advance\@mparbottom\dp\@marbox
    \global\advance\@mparbottom\marginparpush
}
{%
    \global\amir@activeMparBottom\@pageht
    \global\advance\amir@activeMparBottom\@tempdima
    \global\advance\amir@activeMparBottom\dp\@marbox
    \global\advance\amir@activeMparBottom\marginparpush
}
{}
{%
  \PackageError{feedback_tools}
    {Could not patch margin-note stack update}
    {The LaTeX marginpar implementation may have changed.}%
}

\patchcmd{\@opcol}
{\global \@mparbottom \z@ \global \@textfloatsheight \z@}
{%
  \global \@mparbottom \z@
  \global \amir@leftMparBottom \z@
  \global \amir@rightMparBottom \z@
  \global \@textfloatsheight \z@
}
{}
{%
  \PackageError{feedback_tools}
    {Could not patch margin-note page reset}
    {The LaTeX output routine may have changed.}%
}

\renewcommand{\@todonotes@drawMarginNoteWithLine}{%
    \marginpar{%
        \begin{tikzpicture}[remember picture, overlay]%
            \node [coordinate] (inMargin) at (0,0) {};%
        \end{tikzpicture}%
        \@todonotes@drawMarginNote%
        \begin{tikzpicture}[remember picture, overlay]%
          \begin{pgfonlayer}{amirbackground}%
            \draw[amirConnector]
                let
                    \p1 = (amirStart),
                    \p2 = (amirEnd),
                    \p3 = (inMargin)
                in
                \ifdim \x3 < \x1
                    ($(amirStart)-(0.05,0)$)
                        [out=175, in=5]
                        to
                        ($(inMargin)+(\amirNoteWidth,-0.02cm)$)
                \else
                    ($(amirEnd)+(0.05,0)$)
                        [out=5, in=175]
                        to
                        ($(inMargin)+(0,-0.02cm)$)
                \fi;
          \end{pgfonlayer}%
        \end{tikzpicture}%
    }%
    \ifnum\amirDesiredSide=1\relax
        \global\count\@currbox=-2\relax
    \else\ifnum\amirDesiredSide=2\relax
        \global\count\@currbox=-3\relax
    \fi\fi
}

\makeatother

\newcommand{\amir}[3][]{%
    \stepcounter{amirNoteSide}%
    \begin{tikzpicture}[remember picture, overlay, baseline=-0.75ex]%
        \node [coordinate] (amirStart) {};%
    \end{tikzpicture}%
    \hl{#2}%
    \begin{tikzpicture}[remember picture, overlay, baseline=-0.75ex]%
        \node [coordinate] (amirEnd) {};%
    \end{tikzpicture}%
    \ifodd\value{amirNoteSide}%
        \global\amirDesiredSide=1\relax
    \else
        \global\amirDesiredSide=2\relax
    \fi
    \todo[
        size=\tiny,
        color=amirBox,
        bordercolor=amirBorder,
        textcolor=amirText,
        #1
    ]{\textbf{Amir:} #3}%
    \global\amirDesiredSide=0\relax
}

\extrafloats{100}

\usepackage{marginnote}

\makeatletter

\def\amir@captionLeftPrev{}
\def\amir@captionRightPrev{}

\newcommand{\resetamircaptionstack}{%
    \global\let\amir@captionLeftPrev\@empty
    \global\let\amir@captionRightPrev\@empty
}

\DeclareRobustCommand{\amircaption}[3][]{%
    \stepcounter{amirNoteSide}%
    \edef\amir@capid{\theamirNoteSide}%
    \def\amir@captionvshift{0pt}%
    \pgfkeys{%
        /amircaption/.cd,
        vshift/.store in=\amir@captionvshift,
        .unknown/.code={}%
    }%
    \pgfkeys{/amircaption/.cd,#1}%
    \if@twoside
        \ifodd\value{page}%
            \@tempdima=\oddsidemargin
        \else
            \@tempdima=\evensidemargin
        \fi
    \else
        \@tempdima=\oddsidemargin
    \fi
    \advance\@tempdima by 1in
    \advance\@tempdima by \hoffset
    \edef\amir@textleft{\the\@tempdima}%
    \advance\@tempdima by \textwidth
    \edef\amir@textright{\the\@tempdima}%
    \tikz[
        remember picture,
        overlay,
        baseline=-0.75ex
    ]{%
        \coordinate
            (amirCaptionStart-\amir@capid);%
    }%
    \hl{#2}%
    \tikz[
        remember picture,
        overlay,
        baseline=-0.75ex
    ]{%
        \coordinate
            (amirCaptionEnd-\amir@capid);%
    }%
    \tikz[
        remember picture,
        overlay
    ]{%
        \ifodd\value{amirNoteSide}%
            \coordinate
                (amirCaptionDesired-\amir@capid)
                at
                ([xshift=\dimexpr
                    \amir@textleft-\marginparsep
                 \relax,
                  yshift=\amir@captionvshift]
                 current page.south west
                 |- amirCaptionStart-\amir@capid);%
            \ifx\amir@captionLeftPrev\@empty
                \coordinate
                    (amirCaptionPos-\amir@capid)
                    at
                    (amirCaptionDesired-\amir@capid);%
            \else
                \def\amir@captionchoice{desired}%
                \path
                    let
                        \p1 =
                            (amirCaptionDesired-\amir@capid),
                        \p2 =
                            (\amir@captionLeftPrev.south)
                    in
                    \pgfextra{%
                        \ifdim
                            \y1 >
                            \dimexpr
                                \y2-\marginparpush
                            \relax
                            \gdef\amir@captionchoice{previous}%
                        \else
                            \gdef\amir@captionchoice{desired}%
                        \fi
                    };%
                \ifstrequal
                    {\amir@captionchoice}
                    {previous}
                    {%
                        \coordinate
                            (amirCaptionPos-\amir@capid)
                            at
                            ([yshift=-\marginparpush]
                             \amir@captionLeftPrev.south);%
                    }
                    {%
                        \coordinate
                            (amirCaptionPos-\amir@capid)
                            at
                            (amirCaptionDesired-\amir@capid);%
                    }%
            \fi
            \node[
                anchor=north east,
                draw=amirBorder,
                fill=amirBox,
                text=amirText,
                text width=
                    \dimexpr\amirNoteWidth-8pt\relax,
                inner sep=3pt,
                outer sep=0pt,
                font=\tiny,
                align=left
            ]
            (amirCaptionBox-\amir@capid)
            at
            (amirCaptionPos-\amir@capid)
            {%
                \textbf{Amir:} #3%
            };%
            \begin{pgfonlayer}{amirbackground}%
                \draw[
                    amirConnector
                ]
                    ($(amirCaptionStart-\amir@capid)
                        -(0.05,0)$)
                    [out=175,in=5]
                    to
                    (amirCaptionBox-\amir@capid.east);%
            \end{pgfonlayer}%
            \xdef\amir@captionLeftPrev{%
                amirCaptionBox-\amir@capid
            }%
        \else
            \coordinate
                (amirCaptionDesired-\amir@capid)
                at
                ([xshift=\dimexpr
                    \amir@textright+\marginparsep
                 \relax,
                  yshift=\amir@captionvshift]
                 current page.south west
                 |- amirCaptionEnd-\amir@capid);%
            \ifx\amir@captionRightPrev\@empty
                \coordinate
                    (amirCaptionPos-\amir@capid)
                    at
                    (amirCaptionDesired-\amir@capid);%
            \else
                \def\amir@captionchoice{desired}%
                \path
                    let
                        \p1 =
                            (amirCaptionDesired-\amir@capid),
                        \p2 =
                            (\amir@captionRightPrev.south)
                    in
                    \pgfextra{%
                        \ifdim
                            \y1 >
                            \dimexpr
                                \y2-\marginparpush
                            \relax
                            \gdef\amir@captionchoice{previous}%
                        \else
                            \gdef\amir@captionchoice{desired}%
                        \fi
                    };%
                \ifstrequal
                    {\amir@captionchoice}
                    {previous}
                    {%
                        \coordinate
                            (amirCaptionPos-\amir@capid)
                            at
                            ([yshift=-\marginparpush]
                             \amir@captionRightPrev.south);%
                    }
                    {%
                        \coordinate
                            (amirCaptionPos-\amir@capid)
                            at
                            (amirCaptionDesired-\amir@capid);%
                    }%
            \fi
            \node[
                anchor=north west,
                draw=amirBorder,
                fill=amirBox,
                text=amirText,
                text width=
                    \dimexpr\amirNoteWidth-8pt\relax,
                inner sep=3pt,
                outer sep=0pt,
                font=\tiny,
                align=left
            ]
            (amirCaptionBox-\amir@capid)
            at
            (amirCaptionPos-\amir@capid)
            {%
                \textbf{Amir:} #3%
            };%
            \begin{pgfonlayer}{amirbackground}%
                \draw[
                    amirConnector
                ]
                    ($(amirCaptionEnd-\amir@capid)
                        +(0.05,0)$)
                    [out=5,in=175]
                    to
                    (amirCaptionBox-\amir@capid.west);%
            \end{pgfonlayer}%
            \xdef\amir@captionRightPrev{%
                amirCaptionBox-\amir@capid
            }%
        \fi
    }%
}

\makeatother

%% file: tmlr_style/math_commands.tex
\usepackage{amsmath,amsfonts,bm}

\def\eqref#1{equation~\ref{#1}}

\def\1{\bm{1}}

\DeclareMathAlphabet{\mathsfit}{\encodingdefault}{\sfdefault}{m}{sl}
\SetMathAlphabet{\mathsfit}{bold}{\encodingdefault}{\sfdefault}{bx}{n}



%% file: sections/Abstract.tex
Algorithmic recourse methods provide counterfactual explanations that inform individuals of the actions required to overturn an unfavorable model decision. Despite rapid methodological progress, principled comparison remains elusive; existing frameworks are often difficult to extend and lack both interoperability and systematic verification that integrated methods faithfully reproduce their originally reported claims. We introduce \RecourseBench, a unified evaluation framework built around three commitments: modularity, reproducibility, and interactivity. The framework decomposes the pipeline into five fully decoupled layers—Data, Preprocessing, Model, Recourse Method, and Evaluation---governed by abstract interfaces and a dynamic registry. Every integrated method is classified into a four-tier reproducibility taxonomy based on artifact availability, followed by a systematic verification of its core empirical claims. We further provide an interactive web interface for flexible, configuration-driven exploration across datasets, model architectures, methods, and evaluations. To our knowledge, \RecourseBench is the first benchmark to explicitly ground recourse evaluation in structured claim verification and mathematically rigorous reproducibility standards, all while featuring the largest collection of state-of-the-art recourse algorithms (\numMethodsFull in total).

%% file: sections/Introduction.tex
\section{Introduction}
\label{sec:intro}

The use of machine learning models has expanded rapidly in high-stakes domains such as healthcare~\citep{khosravi2024artificial}, finance~\citep{dote2025leveraging}, and criminal justice~\citep{nuredin2024impact,kleinberg2018human}. As these systems increasingly influence consequential decisions, the need for transparency and actionable feedback has become more pronounced. While early Explainable AI (XAI) methods aim to interpret model behavior~\citep{dwivedi2023explainable}, they primarily provide \emph{descriptive} insights and often fail to offer concrete guidance to affected individuals.

Algorithmic recourse addresses this limitation by generating counterfactual explanations that specify how an individual can modify their features to obtain a desired outcome~\citep{wachter_counterfactual_2017,karimi_survey_2022}. In contrast to purely descriptive explanations, recourse methods provide \emph{prescriptive} recommendations, making them particularly relevant in situations where users require explicit guidance to alter predicted outcomes. Moreover, emerging regulatory frameworks such as the European Union’s AI Act~\citep{EU2024AIAct}, GDPR~\citep{voigt2017eu}, and Canada's Directive on Automated Decision-Making~\citep{canada_directive_adm_2025} have increased the demand for meaningful explanations of automated decisions, further motivating the development of actionable recourse methods.

Since the survey on state-of-the-art recourse methods conducted by~\citet{karimi_survey_2022}, the diversity of these algorithms has continued to grow, increasing the need for robust, fair, and scalable evaluation frameworks.
%
Recourse methods are often optimized for different and sometimes conflicting objectives, such as proximity, sparsity, robustness, and feasibility, making direct comparison non-trivial.
Furthermore, evaluation criteria are inherently context-dependent.
For instance, robustness must be assessed differently depending on whether the focus is data shift~\citep{upadhyay_towards_2021} or model multiplicity~\citep{hamman_robust_2024}.
This variability necessitates flexible benchmarking pipelines that can accommodate evolving metrics and experiment settings while maintaining consistent experimental conditions.
A standardized benchmarking framework is therefore essential for ensuring reproducibility, enabling fair comparisons, and allowing practitioners to identify methods that best align with application-specific constraints.

Although the research community has recognized this need, existing open-source solutions leave critical gaps in structural modularity and rigorous verification of the implemented methods.
Historically, many libraries have positioned recourse merely as a sub-feature within broader explainability toolkits (e.g., Alibi~\citep{alibi2021}) or specialized recourse method repositories (e.g., RobustX~\citep{jiang2025robustx}, CounterfactualExplanations.jl~\citep{Altmeyer2023}).
While useful for standalone generation, these tools lack the unified pipelines and diverse datasets required for systematic benchmarking.
%
More recently, dedicated recourse benchmarking frameworks have emerged, but they introduce their own architectural limitations.
%
CARLA~\citep{pawelczyk2021carla} pioneered unified environments for tabular recourse, yet its tightly coupled architecture restricts the seamless integration of new evaluation metrics or model types.
Conversely, ReLax~\citep{Guo2024} achieves high computational efficiency via hardware-accelerated JAX~\citep{jax2018github}, but this design imposes strict ecosystem constraints, complicating integration of the method using other libraries like PyTorch and scikit-learn. Ultimately, current ecosystems leave a critical void for a framework that is simultaneously modular, reproducible, and interactive.


\paragraph{Our Contributions:} We introduce \RecourseBench, a unified benchmarking framework designed around three core principles: modularity, reproducibility, and interactivity. The framework decomposes the benchmarking pipeline into five decoupled layers—Data, Preprocessing, Model, Recourse Method, and Evaluation which enables seamless integration of new components without modifying existing ones. 
The system natively supports widely used machine learning ecosystems, including PyTorch and Scikit-learn. 
Also, we introduce a structured reproducibility taxonomy system that categorizes methods in terms of the reproduction that can be achieved based on existing artifacts. We then validate the scientific claims of each integrated method through automated testing of reported experiments using mathematically sound quantitative assessment techniques, promoting transparency and reliability in benchmarking.
Finally, we present the findings from a comprehensive suite of benchmarking experiments. These include baseline evaluations across four standard datasets and predictive models prevalent in the algorithmic recourse literature, alongside stability analyses and specialized benchmarks targeting sub-domains such as robustness and diversity. To make these findings accessible, we introduce an interactive web interface for exploring the data. This system enables users to construct configurable leaderboards, conduct multi-metric comparisons, and filter methods based on their specific capabilities.

The remainder of this paper is organized as follows. \Cref{sec:prelim} presents the necessary preliminaries on algorithmic recourse and reproducibility. \Cref{sec:benchmarking} describes the architecture of the proposed framework. \Cref{sec:reproducibility} introduces the reproducibility protocol with our three-level categorization and assessment scheme.~\Cref{sec:evaluation} details the evaluation methodology and our benchmarking results. Finally,~\Cref{sec:discussion} provides an analysis of the results with a brief conversation on future work, and~\Cref{sec:conclusion} concludes the paper.

%% file: sections/Preliminaries.tex
\section{Preliminaries}
\label{sec:prelim}

To contextualize our framework, we provide a brief overview of the algorithmic recourse methods and discuss the challenges of reproducibility in machine learning systems.
%

\subsection{Recourse Methods}
\label{recourse_methods}

We organize the recourse methods implemented in our framework into six broad categories.
This categorization reflects common methodological paradigms observed in the literature. 
However, we acknowledge that this classification scheme is representative of our benchmarking suite rather than an exhaustive list of all possible recourse methodologies. Our categorization is an extension of the past work surveying these methods, primarily by~\citet{karimi_survey_2022}.

\textbf{Constraint $\&$ Logic-Based Methods:} This family formulates recourse as a formal constraint satisfaction problem. Methods may employ general-purpose solvers like Mixed-Integer Programming (MIP) and Satisfiability (SAT) \citep{bui2025coverage, ustun_actionable_2019}, or directly exploit the architecture of glass-box models by extracting and satisfying logical root-to-leaf paths in decision trees \citep{tolomei2017interpretable}. While providing guarantees for optimal solutions and strict adherence to complex feasibility rules, these methods often face exponential scaling bottlenecks on large datasets or deep models \citep{marzari2025probabilistically}.

\textbf{Gradient-Based Optimization:} Pioneered by~\citet{wachter_counterfactual_2017} and expanded in subsequent works \citep{upadhyay_towards_2021, pawelczyk_probabilistically_2023}, these approaches treat counterfactual generation as a continuous minimization task. They leverage backpropagation to iteratively perturb inputs across decision boundaries \citep{mothilal_explaining_2020, black2021consistent}. Though computationally efficient and highly scalable, they risk converging on practically infeasible ``off-manifold'' states and require strict regularization to ensure actionability.

\textbf{Representation Learning:} These methods shift the search from the ambient feature space to a learned, lower-dimensional latent manifold via encoder-decoder architectures \citep{downs2020cruds, joshi2019towards, pawelczyk2020learning, mahajan2019preserving}. This generative constraint naturally encourages realistic, distributionally plausible counterfactuals \citep{pegios2025, Wang2025, samoilescu2021}. However, their success is heavily bottlenecked by the quality of the learned representation and the difficulty of decoding discrete features faithfully.

\textbf{Surrogate-Assisted Optimization:} To bypass the opacity and non-convexity of black-box models, these techniques fit a localized approximation around the factual instance. The surrogate can be a deterministic linear hyperplane \citep{kayastha_learning-augmented_2026} or a probabilistic density landscape \citep{nguyen2022robust}. The counterfactual is then found by optimizing directly over this localized surrogate. While highly scalable, the recourse's validity depends entirely on the local fidelity of the surrogate approximation.

\textbf{Derivative-Free $\&$ Heuristic Search:} Treating the underlying model as a strict black box, these methods bypass gradients and formal solvers entirely. Instead, they utilize iterative meta-heuristics to perturb discrete feature combinations \citep{yadav_low-cost_2022} or generate random, synthetic observations in the continuous space to find boundary-crossing points \citep{laugel2017inverse}. This paradigm easily handles non-differentiable models, but lacks global optimality guarantees and can suffer from poor sample efficiency due to high query counts.

\textbf{Instance-Based Search:} Anchoring the search entirely to existing observational data, these methods filter the training dataset for candidate points already in the target class and interpolate geometric paths toward them \citep{poyiadzi2020face, leofante2024promoting}. This data-driven approach naturally avoids unrealistic synthetic states by relying on historical distributions, though its success is strictly limited by the density and diversity of the underlying dataset.

\subsection{Reproducibility in Machine Learning and Software Engineering}
\label{sec:reproducibility_related}

Reproducibility---the ability of an independent research team to obtain equivalent results using the same method and sufficient documentation~\citep{gundersen_state_2018}---is a fundamental requirement for scientific progress.
It is distinct from \emph{replicability}, which refers to reaching similar conclusions under different data or experimental conditions~\citep{sciences_reproducibility_2019}.

While the broader scientific community has long contended with widespread reproducibility failures~\citep{baker_1500_2016}, machine learning (ML) presents a uniquely challenging environment due to the multiplicity of factors that can silently invalidate a result. 
Empirical studies demonstrate that a large proportion of published ML results cannot be reliably reproduced due to incomplete experimental reporting~\citep{gundersen_state_2018, gundersen_machine_2022}. 
Even when source code is available, missing documentation and implicit assumptions frequently prevent faithful reproduction. Recognition of this crisis has prompted strong community-level responses, most notably the establishment of workshops and conference tracks, a prominent one being the Machine Learning Reproducibility Challenge (MLRC).

~\citet{henderson_deep_2018} provide a particularly instructive demonstration of the problem: in deep reinforcement learning, non-determinism in standard benchmark environments, combined with intrinsic algorithmic variance, can render published comparisons statistically uninterpretable. 
Once experimental variance is properly characterized, seemingly distinct algorithms may exhibit no meaningful performance difference---a finding that calls into question the validity of entire lines of comparative work. 
More broadly,~\citet{semmelrock_reproducibility_2025} identify a compounding set of factors inherent to ML research that can shift not merely individual numerical outcomes, but a study's comparative conclusions altogether. 
These difficulties are further amplified by structural incentives within the research community~\citep{desai_what_2025}.

In response to these challenges, recent works have begun looking toward software engineering for solutions. The discipline of regression testing---systematically re-executing a validated test suite to detect behavioral regressions as a codebase evolves~\citep{rothermel_analyzing_1996}---provides a principled, well-established mechanism for enforcing consistency guarantees. 
Drawing explicitly on this tradition, the community increasingly recognizes that maintaining automated, executable test suites is essential for ensuring ML systems and benchmarks can be reliably and externally verified~\citep{rothermel_analyzing_1996}.

Despite these advances, the uptake of such practices within specialized benchmarking repositories remains limited. 
In the domain of algorithmic recourse specifically, existing frameworks such as CARLA~\citep{pawelczyk2021carla} and ReLax~\citep{Guo2024} integrate methods in good faith but provide no
systematic mechanism to validate that their implementations remain faithful to the empirical results originally reported. 
To our knowledge, no prior recourse benchmark explicitly enforces method-level reproducibility through automated quantitative testing.
This gap is consequential: a benchmarking framework can serve as a reliable basis for comparative evaluation only if the fidelity of each constituent method to its original specification has been established.
How we address this gap in the design of our framework is described in~\Cref{sec:reproducibility}.

%% file: sections/Framework.tex
\begin{figure}[htbp!] \centering  
\includegraphics[width=\textwidth]{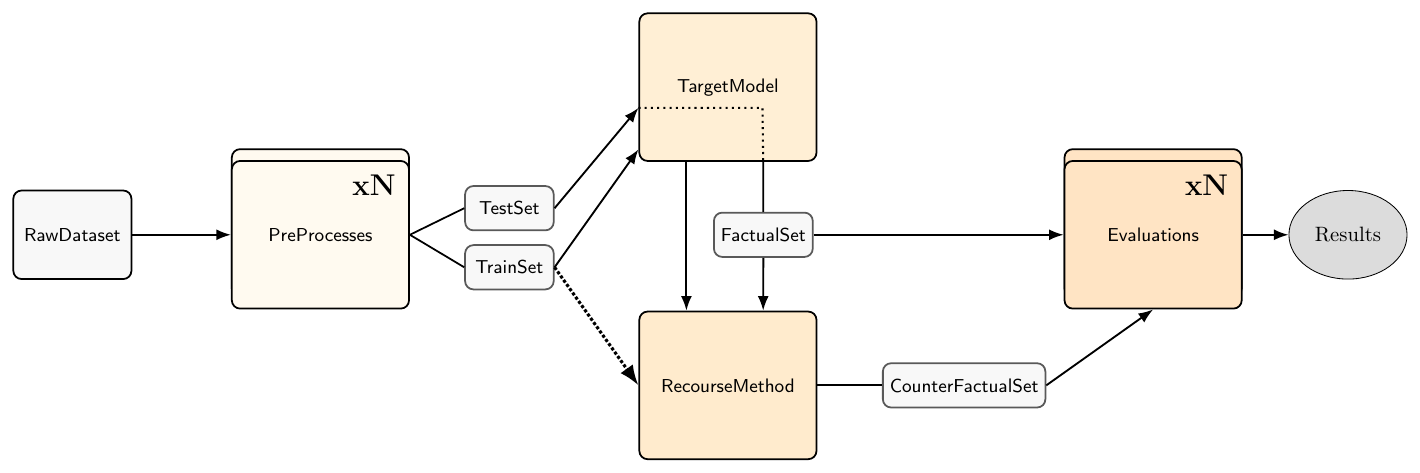} 
\caption{This is a graph of the framework's general workflow. Noted that \texttt{RawDataset} / \texttt{TrainSet} / \texttt{TestSet} / \texttt{FactualSet} / \texttt{CounterfactualSet} are all inherited from \texttt{DatasetObject}. \texttt{TrainSet} is optional for some recourse methods. \texttt{FactualSet} is the target model prediction for \texttt{TestSet}. 
See \autoref{sec:architecture} for more details.}
\label{fig:workflow} 
\end{figure}

\section{Framework Design}
\label{sec:benchmarking}

Our framework addresses the systemic challenges of benchmarking algorithmic recourse by providing a flexible, rigorous environment that accommodates novel concepts, data processing techniques, and evaluation paradigms. To support large-scale collaborative development without compromising scientific rigor, the system adheres to three core design principles:

\textbf{Modularity and Security.} The framework is strictly compartmentalized into independent subsystems (e.g., Dataset, PreProcess, TargetModel, Method, Evaluation) that interact exclusively through predefined public interfaces. To prevent abstraction leaks, messages passed between subsystems are encapsulated in \texttt{DatasetObjects}, making tabular data self-contained. Furthermore, we enforce strict privilege controls; for example, \texttt{TargetModel} components are exposed to \texttt{Method} components as read-only to prevent unintended mutations and cascading side effects. We also utilize an additive development paradigm where new functionality requires only an isolated subfolder inheriting from \texttt{MethodObject} (implementing \texttt{fit()} and \texttt{get\_counterfactuals()}) and a simple registration. Consequently, components act as composable, independently substitutable units that researchers can easily combine (see \autoref{sec:architecture} and \autoref{fig:code} for architectural rationales and minimal code signatures).

\textbf{Reproducibility and Standardization.} To guarantee a level playing field, all methods are subjected to identical, immutable settings orchestrated through a centralized \texttt{Experiment} workflow and governed by a YAML configuration. Randomness is tightly regulated via a dedicated context manager that resets the local seed and pseudo-random sequence, ensuring consistent results even if module execution order changes. The framework also supports the integration of distinct reproduction codebases, allowing isolated, single-file implementations to retain their original architectural idiosyncrasies without compromising the overarching design.

\textbf{Interactivity.} Designed as an active development environment rather than a static leaderboard, \RecourseBench exposes its capabilities through two primary avenues. First, a programmatic PyPI API allows researchers to effortlessly load standardized datasets, models, and baselines into custom Python scripts, eliminating the friction of adapting third-party code. Second, an interactive web dashboard provides rapid, no-code exploration of all pre-computed results. Users can dynamically filter by dataset and architecture, apply preset metric profiles (e.g., \emph{Balanced}, \emph{Quality}, \emph{Speed}), and instantly visualize method trade-offs.

The general workflow of this framework is illustrated in \autoref{fig:workflow}. A visual demonstration of the interface is provided in \Cref{fig:ui}.

%% file: sections/Reproduction.tex
\section{Reproducibility Protocol}
\label{sec:reproducibility}

While the proposed architecture enables flexible and consistent benchmarking, ensuring the reliability of its outputs depends on the faithful reproduction of the integrated methods. In practice, this is often difficult due to incomplete experimental specifications, unavailable code, and sensitivity to implementation details. 

Consequently, there is a critical need for a structured methodology that not only conveys confidence in the implementation of foreign algorithms but also clearly defines what constitutes a successful reproduction. To address this, we introduce a comprehensive, three-pillared methodology for assessing reproducibility: experimental categorization, claim-based verification, and continuous-valued quantitative assessment. A summary of the algorithmic recourse methods is provided in~\Cref{table:technical_summary}, along with detailed per-method reproduction logs in~\Cref{sec:repro_logs}.

\subsection{Categorizing the Evaluation Scope}

Before running any empirical evaluation, it is necessary to define the scope of reproduction that is actually feasible given the available artifacts. We adopt the reproducibility taxonomy established by \citet{pineau2021improving}, which categorizes evaluation efforts based on the availability of the original data and the original code or analysis.

Under this framework, every recourse method integrated into our benchmark is first classified into one of four quadrants: (1) \textbf{Reproducible:} Both the original code and original data are available and utilized; (2) \textbf{Replicable:} The original code is available, but the experiments are run on different data; (3) \textbf{Robust:} The original data is used, but the code and analysis are independently implemented; and (4) \textbf{Generalizable:} Both the code and data differ from the original study, testing the broader theoretical applicability of the algorithm.

A visualization of this taxonomy is provided in \Cref{fig:pineau_matrix}. This initial categorization clarifies the reproduction effort, ensuring that methods evaluated without available code or data are assessed fairly on their generalizability rather than penalized for a lack of exact reproducibility.

\input{sections/repro}

\subsection{Claim-Based Verification}

Drawing on established practices from the Machine Learning Reproducibility Challenge (MLRC) 2022, we observe that successful reproduction efforts often prioritize verifying the central claims of a paper rather than strictly replicating isolated experiments. 

Borrowing from this paradigm, our framework shifts the evaluative focus toward verifying the core theoretical and empirical claims made by the original authors. For every evaluated method, we:
\begin{itemize}
    \item Explicitly extract the primary claims made in the original work alongside their supporting experiments.
    \item Execute the method within our standardized environment to determine if the observed behavior supports these fundamental claims.
\end{itemize}

The extracted claims generally fall into two categories: quantitative and qualitative. A quantitative claim might involve a reported scalar metric demonstrating improved algorithmic sparsity, which we can directly test and compare. Conversely, a qualitative claim, such as a mathematically derived proof of optimality, is not empirically reproducible within a programmatic framework and is therefore excluded from this specific testing phase.

This structure provides a more robust foundation for assessing algorithmic validity beyond exact numerical matching. For instance, an independent implementation may fail to hit the exact numerical targets of its original publication due to underlying framework differences. However, if it successfully demonstrates its core claim---such as generating counterfactual explanations without requiring internal model access---the claim is considered verified/partially verified.

\subsection{Quantitative Assessment}

While claim verification captures the qualitative intent of a method, we still require a mathematically sound mechanism to measure scalar value discrepancies across comparable experiments. To achieve this, we integrate the Extended Quantified Reproducibility Assessment (QRA++) framework \citep{belz2025qra++}, which replaces binary success thresholds with continuous-valued reproducibility assessments.

For standard single numerical scores, such as validity or sparsity metrics, we utilize the unbiased small-sample coefficient of variation ($CV^*$)to measure measurement precision. In machine learning benchmarking, sample sizes for reproduction are often small, which can cause standard coefficient of variation metrics to yield overly optimistic results. The $CV^*$ metric corrects for this bias and is defined as follows:

\begin{equation}
    CV^* = \left(1+\frac{1}{4n}\right)\frac{s^*}{|m|}
\end{equation}

In this formulation, $n$ represents the sample size, $m$ is the sample mean, and $s^{*}$ is the unbiased sample standard deviation. By adopting this literature-backed measure, \RecourseBench evaluates reproducibility in a manner that is continuous, statistically grounded, and directly comparable across different studies and methods. The results of the reproduction study, along with the $CV^*$ measurements, are presented in~\Cref{sec:repro_logs}.

\input{sections/table}

%% file: sections/repro.tex
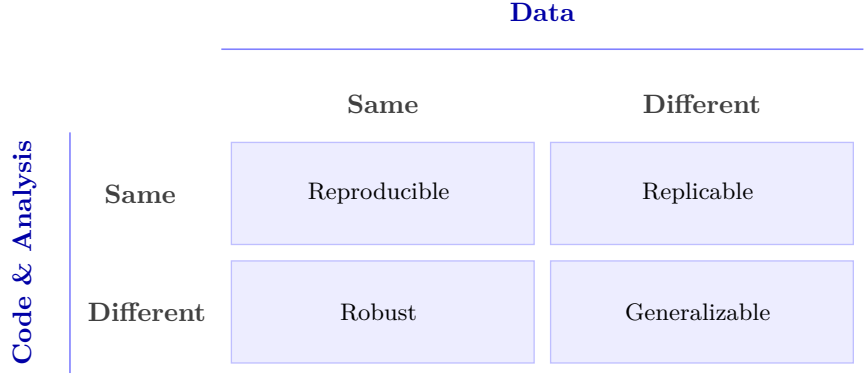
\begin{figure}[t]
\centering

\begin{tikzpicture}[
    font=\small,
    cell/.style={
        minimum width=4.0cm,
        minimum height=1.35cm,
        align=center,
        draw=blue!25,
        fill=blue!7,
        line width=0.5pt
    },
    hdr/.style={
        font=\bfseries,
        text=black!75
    },
    axislabel/.style={
        font=\bfseries,
        text=blue!65!black
    }
]

\matrix (m) [
    matrix of nodes,
    nodes={cell},
    row sep=2mm,
    column sep=2mm
] {
    Reproducible & Replicable    \\
    Robust       & Generalizable \\
};

\node[hdr] at ([yshift=5mm]m-1-1.north) {Same};
\node[hdr] at ([yshift=5mm]m-1-2.north) {Different};

\node[hdr, anchor=east] at ([xshift=-6mm]m-1-1.west) {Same};
\node[hdr, anchor=east] at ([xshift=-2mm]m-2-1.west) {Different};

\draw[blue!50, line width=0.6pt] 
    ([yshift=11mm]m.north west) -- ([yshift=11mm]m.north east);

\node[axislabel] at ([yshift=16mm]m.north) {Data};

\draw[blue!50, line width=0.6pt] 
    ([xshift=-20mm]m.north west) -- ([xshift=-20mm]m.south west);

\node[axislabel, rotate=90] at ([xshift=-26mm]m.west) {Code \& Analysis};

\end{tikzpicture}

\caption{
The reproducibility taxonomy adapted from \cite{pineau2021improving},
illustrating four evaluation settings according to whether the
code and analysis, and data are held fixed or changed.
}
\label{fig:pineau_matrix}

\end{figure}

%% file: sections/table.tex
\newcommand{\algcite}[2]{#1~{\scriptsize\citep{#2}}}

\newcommand{\yy}{\fullcircblack}
\newcommand{\ex}{\fullcircgray}
\newcommand{\pp}{\halfcircblack}
\newcommand{\xx}{\emptycircblack}
\newcommand{\qq}{?}

\newcommand{\testoptimal}{{\footnotesize \faStar}}
\newcommand{\testcoverage}{{\footnotesize \faSignal}}
\newcommand{\testrealtime}{{\footnotesize \faClockO}}

\newcommand*\fullcircblack[1][.9ex]{\tikz\fill (0,0) circle (#1);}
\newcommand*\fullcircgray[1][.93ex]{\tikz\fill[gray!75] (0,0) circle (#1);}
\newcommand*\halfcircblack[1][.88ex]{%
  \begin{tikzpicture}
    \draw[fill][gray!75] (0,0)-- (90:#1) arc (90:270:#1) -- cycle ;
    \draw (0,0) circle (#1);
  \end{tikzpicture}
}
\newcommand*\emptycircblack[1][.9ex]{\tikz\draw[gray!75] (0,0) circle (#1);}

\renewcommand{\arraystretch}{0.98}


\begingroup

\setlength{\tabcolsep}{2pt}
\renewcommand{\arraystretch}{1.15}

\scriptsize

\setlength{\LTleft}{0pt}
\setlength{\LTright}{0pt}

\begin{longtable}{
  @{} >{\raggedright\arraybackslash}p{4.8cm}
  c c c c
  c c
  c c c
  c c
  >{\raggedright\arraybackslash}p{1.7cm}
  >{\raggedright\arraybackslash}p{1.7cm}
  @{}
}
  \caption{ 
    An overview of recourse algorithms for consequential decision-making settings is presented.
    Ordered chronologically, we summarize the \emph{goal}, \emph{formulation}, and \emph{solution} of each algorithm.
    Symbols are used to indicate supported settings in the experimental section of their paper (\yy), settings that are natural extensions of the presented algorithm (\ex), settings that are partially supported (\pp\!\!), and settings that are not supported (\xx).
    The models cover a broad range of tree-based (TB), kernel-based (KB), differentiable (DF), or other (OT) types.
    Actionability constraints (unconditional or conditional), plausibility constraints (domain-, density-, and prototypical-consistency), and additional constraints (diversity, sparsity) are also explored.
  }
  \label{table:technical_summary}\\

  \toprule

  \multirow{3}{*}{Algorithm}
  & \multicolumn{11}{c}{Formulation}
  & \multicolumn{2}{c}{Solution} \\

  \cmidrule(lr){2-12}
  \cmidrule(lr){13-14}

  & \multicolumn{4}{c}{Model}
  & \multicolumn{2}{c}{Actionability}
  & \multicolumn{3}{c}{Plausibility}
  & \multicolumn{2}{c}{Extra}
  & \multicolumn{1}{l}{Type} 
  & \multicolumn{1}{c}{Taxonomy} \\

  \cmidrule(lr){2-5}
  \cmidrule(lr){6-7}
  \cmidrule(lr){8-10}
  \cmidrule(lr){11-12}

  & TB & KB & DF & OT
  & uncond. & cond.
  & dom. & dens. & proto.
  & diver. & spar.
  & \\

  \midrule

  \endfirsthead

  \midrule

  \endhead

  \midrule
  \multicolumn{13}{r}{\footnotesize Continued on next page}\\
  \midrule
  \endfoot

  \bottomrule
  \endlastfoot

  \algcite{Feature Tweak}{tolomei2017interpretable}
    & \yy & \xx & \xx & \xx
    & \xx & \xx
    & \xx & \xx & \xx
    & \xx & \xx
    & Constraint & Generalizable \\

  Wachter~\citep{wachter_counterfactual_2017}
    & \xx & \xx & \yy & \xx
    & \yy & \xx
    & \xx & \xx & \xx
    & \yy & \yy
    & Gradient & Generalizable \\

  GS~\citep{laugel2017inverse}
    & \yy & \yy & \ex & \ex
    & \xx & \xx
    & \xx & \xx & \xx
    & \xx & \yy
    & Heuristic & Reproducible \\

  DiCE~\citep{mothilal_explaining_2020}
    & \xx & \xx & \yy & \xx
    & \yy & \xx
    & \ex & \xx & \xx
    & \yy & \xx
    & Gradient & Reproducible \\


  REVISE~\citep{joshi2019towards}
    & \xx & \xx & \yy & \xx
    & \yy & \xx
    & \yy & \yy & \xx
    & \xx & \xx
    & Represent. & Generalizable \\

  FACE~\citep{poyiadzi2020face}
    & \ex & \ex & \yy & \ex
    & \yy & \yy
    & \ex & \yy & \yy
    & \xx & \xx
    & Inst. Search & Generalizable \\

  C-CHVAE~\citep{pawelczyk2020learning}
    & \ex & \yy & \yy & \ex
    & \yy & \xx
    & \yy & \yy & \xx
    & \xx & \xx
    & Represent. & Reproducible \\

  CFVAE~\citep{mahajan2019preserving}
    & \xx & \xx & \yy & \xx
    & \xx & \xx
    & \pp & \xx & \xx
    & \xx & \yy
    & Represent. & Reproducible \\

  CRUDS~\citep{downs2020cruds}
    & \ex & \ex & \yy & \ex
    & \yy & \yy
    & \xx & \yy & \xx
    & \yy & \xx
    & Represent. & Generalizable \\

  ROAR~\citep{upadhyay_towards_2021}
    & \xx & \xx & \yy & \yy
    & \yy & \xx
    & \xx & \yy & \xx
    & \yy & \yy
    & Gradient & Reproducible \\

  CLUE~\citep{antoran_getting_2021}
    & \xx & \xx & \yy & \xx
    & \xx & \xx
    & \yy & \yy & \xx
    & \pp & \yy
    & Represent. & Reproducible \\

  CFRL~\citep{samoilescu2021}
    & \yy & \xx & \yy & \yy
    & \yy & \yy
    & \yy & \xx & \xx
    & \yy & \yy
    & Represent. & Robust \\

  SNS~\citep{black2021consistent}
    & \xx & \xx & \yy & \pp
    & \xx & \xx
    & \xx & \xx & \xx
    & \xx & \xx
    & Gradient & Robust \\

  EMC-COLS~\citep{yadav_low-cost_2022}
    & \xx & \xx & \yy & \yy
    & \yy & \yy
    & \xx & \xx & \xx
    & \yy & \yy
    & Heuristic & Reproducible \\

  RBR~\citep{nguyen2022robust}
    & \xx & \xx & \yy & \yy
    & \yy & \xx
    & \yy & \yy & \xx
    & \xx & \yy
    & Surrogate-Assisted & Reproducible \\

  CoGS~\citep{virgolin_robustness_2023}
    & \xx & \xx & \yy & \xx
    & \yy & \yy
    & \xx & \xx & \xx
    & \xx & \xx
    & Constraint & Reproducible \\

  ClaPROAR~\citep{altmeyer_endogenous_2023}
    & \xx & \xx & \yy & \xx
    & \xx & \xx
    & \yy & \xx & \xx
    & \xx & \xx
    & Gradient & Generalizable \\

  Gravitational~\citep{altmeyer_endogenous_2023}
    & \xx & \xx & \yy & \xx
    & \xx & \xx
    & \yy & \xx & \yy
    & \xx & \xx
    & Gradient & Generalizable \\

  Proplace~\citep{jiang_provably_2024}
    & \xx & \xx & \yy & \xx
    & \xx & \xx
    & \yy & \xx & \xx
    & \xx & \xx
    & Constraint & Reproducible \\

  CEMSP~\citep{wang_flexible_2023}
    & \xx & \xx & \yy & \yy
    & \yy & \yy
    & \yy & \xx & \xx
    & \yy & \yy
    & Constraint & Reproducible \\

  PROBE~\citep{pawelczyk_probabilistically_2023}
    & \yy & \xx & \yy & \xx
    & \yy & \xx
    & \xx & \xx & \xx
    & \xx & \yy
    & Gradient & Reproducible \\

  Diverse Dist.~\citep{leofante2024promoting}
    & \xx & \xx & \xx & \xx
    & \xx & \xx
    & \xx & \xx & \yy
    & \xx & \xx
    & Inst. Search & Reproducible \\

  TReX~\citep{hamman_robust_2024}
    & \xx & \xx & \yy & \xx
    & \yy & \xx
    & \yy & \yy & \xx
    & \xx & \yy
    & Gradient & Reproducible \\

  CVAS~\citep{bui2025coverage}
    & \xx & \xx & \yy & \yy
    & \yy & \xx
    & \yy & \yy & \xx
    & \xx & \yy
    & Surrogate-Assisted & Reproducible \\

  ArgEnsemble~\citep{jiang_argumentative_2025}
    & \xx & \xx & \yy & \pp
    & \pp & \xx
    & \xx & \xx & \xx
    & \xx & \pp
    & Constraint & Reproducible \\

  APAS~\citep{marzari2025probabilistically}
    & \xx & \xx & \yy & \xx
    & \xx & \xx
    & \xx & \xx & \xx
    & \xx & \xx
    & Constraint & Reproducible \\

  LARR~\citep{kayastha_learning-augmented_2026}
    & \xx & \xx & \yy & \pp
    & \yy & \xx
    & \xx & \xx & \xx
    & \xx & \yy
    & Surrogate-Assisted & Reproducible \\

\end{longtable}

\endgroup

\normalsize

%% file: sections/Evaluation.tex
\section{Evaluation and Empirical Benchmarking}
\label{sec:evaluation}

A central objective of this work is to provide the research community with empirical insights into the comparative performance of state-of-the-art recourse methods.
To this end, our framework separates three distinct concerns: the evaluation design, which defines the metrics and sampling strategies; the empirical results, which map the performance and trade-offs of the integrated methods; and the benchmark interface, which exposes these results through an easy-to-interact-with visual environment, along with a library for practitioners to use and expand this framework for their own research.

\subsection{Evaluation Metrics and Design}
\label{sec:eval_design}

\paragraph{Evaluation metrics:}
Algorithmic recourse methods optimize for objectives that are inherently in tension, meaning no single metric is sufficient to characterize algorithmic quality. Let $\mathbf{x} \in \mathbb{R}^d$ denote a factual instance, $\mathbf{x}' \in \mathbb{R}^d$ its generated counterfactual, $f$ the target predictive model, $y^*$ the desired class label, and $\{\mathbf{x}_i, \mathbf{x}'_i\}_{i=1}^{N}$ the set of evaluated pairs. We evaluate each method across the following dimensions:

\begin{itemize}[leftmargin=*]
    
    \item 
    \begin{minipage}[c]{0.60\linewidth}
        \textbf{Validity (Success Rate):} The proportion of input instances for which the method produces a counterfactual that achieves the desired class label.
    \end{minipage}\hfill
    \begin{minipage}[c]{0.37\linewidth}
        \setlength{\abovedisplayskip}{0pt}
        \setlength{\belowdisplayskip}{0pt}
        \begin{equation}
            V = \frac{1}{N}\sum_{i=1}^{N}\mathbf{1}\bigl[f(\mathbf{x}'_i) = y^*\bigr]
        \end{equation}
    \end{minipage}

    \vspace{2mm} 

    \item 
    \begin{minipage}[c]{0.60\linewidth}
        \textbf{Proximity ($\ell_2$ distance):} The mean Euclidean distance between each factual instance and its counterfactual, capturing the overarching cost of the recommended changes.
    \end{minipage}\hfill
    \begin{minipage}[c]{0.37\linewidth}
        \setlength{\abovedisplayskip}{0pt}
        \setlength{\belowdisplayskip}{0pt}
        \begin{equation}
            P = \frac{1}{N}\sum_{i=1}^{N}\|\mathbf{x}_i - \mathbf{x}'_i\|_2
        \end{equation}
    \end{minipage}

    \vspace{2mm}

    \item 
    \begin{minipage}[c]{0.60\linewidth}
        \textbf{Sparsity ($\ell_0$ distance):} The mean number of features modified. Sparse recourse is preferred as interventions on fewer features are cognitively and operationally simpler.
    \end{minipage}\hfill
    \begin{minipage}[c]{0.37\linewidth}
        \setlength{\abovedisplayskip}{0pt}
        \setlength{\belowdisplayskip}{0pt}
        \begin{equation}
            S = \frac{1}{N}\sum_{i=1}^{N}\|\mathbf{x}_i - \mathbf{x}'_i\|_0
        \end{equation}
    \end{minipage}
    
    \vspace{2mm}

    \item 
    \begin{minipage}[c]{0.60\linewidth}
        \textbf{Plausibility (yNN):} The mean Euclidean distance from each generated counterfactual to its $y$ nearest neighbors in the training set ($\mathbf{z}^{(j)}_i$). A lower value indicates the counterfactual lies closer to the observed data distribution.
    \end{minipage}\hfill
    \begin{minipage}[c]{0.37\linewidth}
        \setlength{\abovedisplayskip}{0pt}
        \setlength{\belowdisplayskip}{0pt}
        \begin{equation}
            \text{yNN} = \frac{1}{N}\sum_{i=1}^{N}\frac{1}{y}\sum_{j=1}^{y}\|\mathbf{x}'_i - \mathbf{z}^{(j)}_i\|_2
        \end{equation}
    \end{minipage}

    \vspace{2mm}

    \item 
    \begin{minipage}[c]{0.60\linewidth}
        \textbf{Runtime:} The total time required to generate counterfactuals ($t_i$), measured on a consistent computational platform.
    \end{minipage}\hfill
    \begin{minipage}[c]{0.37\linewidth}
        \setlength{\abovedisplayskip}{0pt}
        \setlength{\belowdisplayskip}{0pt}
        \begin{equation}
            T = \sum_{i=1}^{N} t_i
        \end{equation}
    \end{minipage}

\end{itemize}

\begin{table}[htbp]
    \centering
    \caption{
        Summary of the benchmarking results.
        These results are averages computed for negatively predicted instances on a linear and mlp model.
        For each method, we measure Validity, $l_0$ (Sparsity), $l_2$ (Proximity), yNN, and Runtime.
        %
        More details about the experiment setup can be found in~\Cref{sec:evaluation}. 
        First- and second-best results are \textbf{bolded} and \underline{underlined}, respectively.
        Missing values(-) indicate a method was incompatible or failed to generate valid counterfactuals.}
    \label{tab:overall_benchmark}
    \begin{subtable}[t]{\textwidth}
    \centering
    \scriptsize
    \caption{COMPAS Dataset}
    \label{tab:compas_combined}
    \resizebox{\textwidth}{!}{%
    \begin{tabular}{lrrrrrrrrrr}
        \toprule
        & \multicolumn{5}{c}{\textbf{Linear Model}} & \multicolumn{5}{c}{\textbf{MLP Model}} \\
        \cmidrule(lr){2-6} \cmidrule(lr){7-11}
        Method & Val $\uparrow$ & L0 $\downarrow$ & L2 $\downarrow$ & yNN-5 $\downarrow$ & Time (s) $\downarrow$ & Val $\uparrow$ & L0 $\downarrow$ & L2 $\downarrow$ & yNN-5 $\downarrow$ & Time (s) $\downarrow$ \\
        \midrule
        APAS & - & - & - & - & - & \textbf{1.000} & 1.160 & 0.213 & 0.010 & 1.419 \\
        ArgEnsemble & \textbf{1.000} & \textbf{1.000} & 0.217 & 0.059 & 0.822 & \textbf{1.000} & 1.140 & 0.222 & \underline{0.005} & 0.801 \\
        C-CHVAE & \textbf{1.000} & 1.460 & 0.609 & 0.146 & 0.833 & \textbf{1.000} & 1.240 & 0.345 & 0.148 & 0.643 \\
        CEMSP & 0.580 & \textbf{1.000} & 0.170 & 0.016 & 0.509 & 0.580 & \underline{1.034} & 0.187 & \textbf{0.003} & 0.530 \\
        CFRL & \textbf{1.000} & \textbf{1.000} & 0.249 & 0.051 & \underline{0.502} & 0.840 & \textbf{1.000} & 0.177 & 0.018 & \underline{0.501} \\
        CFVAE & \textbf{1.000} & 3.880 & 1.667 & \textbf{0.003} & 0.549 & 0.380 & 3.474 & 1.527 & 0.008 & \textbf{0.454} \\
        ClaPROAR & \textbf{1.000} & 7.000 & 0.446 & 0.153 & 0.641 & 0.780 & 7.000 & 0.200 & 0.097 & 0.681 \\
        CoGS & \textbf{1.000} & \textbf{1.000} & 0.169 & 0.031 & 1.202 & \textbf{1.000} & 1.120 & 0.216 & 0.010 & 1.198 \\
        EMC-COLS & \textbf{1.000} & \textbf{1.000} & 0.169 & 0.027 & 96.909 & \textbf{1.000} & 1.080 & 0.246 & 0.034 & 96.164 \\
        CRUDS & 0.380 & 1.421 & 0.493 & \underline{0.009} & 3.332 & 0.560 & 1.607 & 0.661 & 0.008 & 3.825 \\
        CVAS & \textbf{1.000} & \textbf{1.000} & 0.296 & 0.059 & 1.584 & 0.880 & \textbf{1.000} & 0.235 & 0.020 & 1.458 \\
        DiCE & \textbf{1.000} & \textbf{1.000} & 0.348 & 0.073 & 264.735 & \textbf{1.000} & 1.220 & 0.438 & 0.022 & 255.195 \\
        Diverse Dist. & - & - & - & - & - & \textbf{1.000} & 1.140 & 0.191 & 0.026 & 0.541 \\
        FACE & \underline{0.960} & \textbf{1.000} & 0.220 & 0.046 & 1.318 & 0.840 & \textbf{1.000} & \underline{0.073} & \underline{0.005} & 1.332 \\
        Gravitational & 0.100 & \textbf{1.000} & \textbf{0.119} & 0.027 & \textbf{0.443} & 0.800 & 1.375 & 0.444 & 0.007 & 0.602 \\
        GS & \textbf{1.000} & \underline{1.400} & 0.490 & 0.022 & 0.957 & \textbf{1.000} & 1.260 & 0.386 & 0.009 & 1.361 \\
        LARR & \textbf{1.000} & 4.720 & 0.756 & 0.297 & 0.524 & 0.000 & - & - & - & 0.565 \\
        PROBE & \textbf{1.000} & 7.000 & 0.375 & 0.344 & 15.989 & 0.680 & 2.059 & 0.089 & 0.045 & 675.208 \\
        Proplace & \textbf{1.000} & \textbf{1.000} & 0.178 & 0.033 & 3.052 & \textbf{1.000} & 1.180 & 0.303 & 0.012 & 37.840 \\
        RBR & \textbf{1.000} & 7.000 & 0.316 & 0.123 & 236.136 & \textbf{1.000} & 7.000 & 0.262 & 0.133 & 280.185 \\
        Revise & \textbf{1.000} & \textbf{1.000} & 0.558 & 0.160 & 3.222 & 0.700 & \textbf{1.000} & 0.330 & 0.054 & 3.674 \\
        ROAR & \textbf{1.000} & 7.000 & 0.663 & 0.430 & 59.156 & 0.880 & 7.000 & 0.371 & 0.268 & 27.744 \\
        SNS & - & - & - & - & - & 0.860 & 4.953 & 0.199 & 0.059 & 147.590 \\
        TReX & \textbf{1.000} & 3.880 & 0.433 & 0.134 & 17.631 & \underline{0.960} & 4.667 & 0.458 & 0.176 & 23.988 \\
        Wachter & \textbf{1.000} & \textbf{1.000} & \underline{0.162} & 0.027 & 0.889 & 0.740 & \textbf{1.000} & \textbf{0.070} & 0.011 & 40.160 \\
        \bottomrule
        \end{tabular}%
    }
    \end{subtable}

    \vspace{0.3cm}
    \begin{subtable}[t]{\textwidth}
    \centering
    \scriptsize
    \caption{Credit Dataset}
    \label{tab:credit_combined}
    \resizebox{\textwidth}{!}{%
    \begin{tabular}{lrrrrrrrrrr}
        \toprule
        & \multicolumn{5}{c}{\textbf{Linear Model}} & \multicolumn{5}{c}{\textbf{MLP Model}} \\
        \cmidrule(lr){2-6} \cmidrule(lr){7-11}
       Method & Val $\uparrow$ & L0 $\downarrow$ & L2 $\downarrow$ & yNN-5 $\downarrow$ & Time (s) $\downarrow$ & Val $\uparrow$ & L0 $\downarrow$ & L2 $\downarrow$ & yNN-5 $\downarrow$ & Time (s) $\downarrow$ \\
        \midrule
        APAS & - & - & - & - & - & \textbf{1.000} & 5.960 & \underline{0.159} & 0.211 & 4.048 \\
        ArgEnsemble & \textbf{1.000} & 5.860 & 0.470 & 0.128 & 1.012 & \textbf{1.000} & 6.440 & 0.514 & 0.153 & 0.965 \\
        C-CHVAE & \textbf{1.000} & 12.180 & 2.118 & 1.037 & 0.779 & \textbf{1.000} & 11.880 & 2.110 & 0.903 & 0.800 \\
        CEMSP & \textbf{1.000} & \underline{1.460} & 0.822 & 0.294 & 1.085 & \textbf{1.000} & 2.200 & 0.919 & 0.234 & 1.082 \\
        CFRL & \textbf{1.000} & 9.340 & 1.062 & 0.199 & 0.683 & 0.960 & 9.312 & 1.316 & 0.155 & 0.666 \\
        CFVAE & \textbf{1.000} & 12.820 & 2.137 & \textbf{0.026} & \textbf{0.707} & \textbf{1.000} & 11.880 & 1.998 & \textbf{0.037} & \underline{0.743} \\
        ClaPROAR & \textbf{1.000} & 20.000 & 0.345 & 0.348 & 0.778 & 0.640 & 20.000 & 0.208 & 0.251 & 0.768 \\
        CoGS & \textbf{1.000} & \textbf{1.000} & 0.331 & 0.335 & 1.462 & \underline{0.980} & \underline{1.306} & 0.416 & 0.387 & 1.502 \\
        EMC-COLS & \textbf{1.000} & 1.620 & 0.254 & 0.243 & 105.351 & \textbf{1.000} & 2.900 & 0.521 & 0.355 & 116.053 \\
        CRUDS & \textbf{1.000} & 11.260 & 1.633 & 0.242 & 3.833 & \textbf{1.000} & 11.040 & 1.566 & 0.248 & 4.293 \\
        CVAS & \underline{0.980} & \textbf{1.000} & 0.347 & 0.361 & 1.839 & 0.820 & \textbf{1.000} & 0.313 & 0.338 & 1.699 \\
        DiCE & \textbf{1.000} & 7.180 & 0.712 & 0.578 & 285.311 & \textbf{1.000} & 7.800 & 1.083 & 0.323 & 348.937 \\
        Diverse Dist. & - & - & - & - & - & \textbf{1.000} & 6.520 & 0.380 & 0.186 & \textbf{0.711} \\
        FACE& 0.860 & 5.930 & 0.387 & \underline{0.112} & 1.135 & 0.900 & 6.356 & 0.470 & 0.148 & 1.219 \\
        Gravitational & \textbf{1.000} & 11.540 & 1.747 & 0.258 & \underline{0.765} & \textbf{1.000} & 11.160 & 1.607 & 0.237 & 0.850 \\
        GS & \textbf{1.000} & 2.520 & 1.496 & 1.063 & 5.868 & \textbf{1.000} & 4.220 & 1.797 & 1.229 & 5.562 \\
        LARR & \textbf{1.000} & 9.960 & 1.154 & 1.148 & 0.799 & 0.780 & 10.513 & 1.374 & \underline{0.098} & 1.462 \\
        PROBE & \textbf{1.000} & 20.000 & \underline{0.199} & 0.235 & 19.122 & 0.500 & 9.000 & \textbf{0.139} & 0.219 & 764.374 \\
        Proplace & \textbf{1.000} & 6.020 & 0.444 & 0.151 & 7.790 & \textbf{1.000} & 6.160 & 0.646 & 0.128 & 91.312 \\
        RBR & \textbf{1.000} & 20.000 & 0.347 & 0.253 & 135.002 & \textbf{1.000} & 20.000 & 0.381 & 0.338 & 238.569 \\
        Revise & 0.940 & 9.000 & 0.358 & 0.272 & 3.536 & 0.580 & 9.000 & 0.309 & 0.243 & 3.887 \\
        ROAR & \textbf{1.000} & 20.000 & 0.542 & 0.522 & 35.690 & \textbf{1.000} & 20.000 & 1.583 & 1.235 & 60.839 \\
        SNS & - & - & - & - & - & 0.760 & 15.395 & 0.256 & 0.299 & 145.538 \\
        TReX & \textbf{1.000} & 14.700 & 0.332 & 0.330 & 17.661 & 0.900 & 14.578 & 0.394 & 0.402 & 24.656 \\
        Wachter & \textbf{1.000} & 7.720 & \textbf{0.128} & 0.170 & 0.806 & 0.700 & 7.886 & 0.165 & 0.213 & 37.722 \\
        \bottomrule
    \end{tabular}%
    }
    \end{subtable}
\end{table}

\paragraph{Data sets:}
All datasets evaluated in our benchmark undergo a standardized preprocessing pipeline comprising binary normalization, min-max scaling, categorical one-hot encoding, and a stratified 80/20 train-test split. For the \textbf{Adult income}~\citep{adult_dataset} classification task, recourse aims for the high-income class; we restrict \textit{Sex} and \textit{Age} as immutable while leaving socioeconomic attributes actionable, yielding a 51-feature encoded representation. The \textbf{COMPAS}~\citep{compas_dataset} recidivism dataset targets the non-recidivating class; demographic attributes (\textit{AgeGroup}, \textit{Race}, \textit{Sex}) are fixed as immutable while criminal history remains mutable, resulting in 7 encoded features. The \textbf{Boston Housing}~\citep{Boston_housing_dataset} dataset is adapted for binary classification and serves as a highly permissive testing environment where all non-target features are actionable, expanding to 21 encoded features. Finally, the \textbf{Credit}~\citep{Credit_dataset} default-risk task seeks the favorable non-default outcome; demographic features (\textit{isMale}, \textit{AgeGroup}) are strictly immutable while financial behaviors remain actionable, resulting in a 20-feature model-facing representation.

\paragraph{Predictive models:}
Our benchmark employs four distinct predictive models. The baseline \textbf{Linear} model consists of a single fully connected PyTorch layer mapping inputs to binary class logits, optimized with Adam (learning rate 0.03, batch size 32) over 30 epochs. For the standard neural network, we utilize an \textbf{MLP} comprising two hidden layers (64 and 32 units) with ReLU activations, trained for 50 epochs with Adam (learning rate 0.01, batch size 32). To support tree-specific recourse algorithms such as Feature Tweak~\citep{tolomei2017interpretable}, we integrate a non-neural Scikit-Learn \textbf{Random Forest} ensemble with 100 trees, unbounded maximum depth, and square-root feature subsampling. Finally, to accommodate uncertainty-aware methods such as CLUE~\citep{antoran_getting_2021}, we include a \textbf{Bayesian MLP} (two 64-unit hidden layers) trained via Stochastic Gradient Hamiltonian Monte Carlo (SGHMC) sampling for 200 epochs (learning rate 0.01, batch size 256), which averages predictions over up to 20 stored posterior parameter samples. The performance results of the models, along with hardware specifications, can be found in~\Cref{sec:further_benchmark} and~\Cref{tab:model_performance}

\paragraph{Experimental coverage and sampling:}
A distinguishing feature of our benchmarking approach is the systematic enumeration of experimental configurations. We evaluate combinations of recourse methods, datasets, and predictive model architectures whose compatibility is declared in the per-method configuration files. In total, the benchmark currently presents the evaluation of 4 datasets, 4 model types, and 27 recourse methods, yielding roughly 200~\footnote{This number does not include experiments with FeatureTweak and Clue} compatible method--model--dataset configurations. The final number of configurations takes into account possible mismatches overall, which are discussed within the results.

To evaluate these configurations, instances are drawn from the held-out test set that received an unfavourable prediction. We cap this sample size at 50 instances uniformly at random. This threshold was selected as a practical ceiling; while larger datasets can support wider sampling, smaller tabular datasets frequently yield fewer than 50 unfavourable test-set instances, making this a necessary upper bound to ensure balanced, comparable evaluation across all dataset scales. We present the results of our experiments on the COMPAS and Credit datasets in~\Cref{tab:overall_benchmark}, along with the Adult and Boston housing datasets and methods Feature Tweak and CLUE results available inside~\Cref{sec:further_benchmark}.

\subsection{Specialized Evaluations: Diversity and Robustness}
\label{sec:specialized_benchmarks}

While standard metrics like proximity and sparsity provide a foundational understanding of algorithmic performance, the rapidly expanding scope of recourse research has introduced highly specialized sub-domains. Two prominent ones are \emph{counterfactual diversity}, which provides users with multiple, distinct pathways to recourse, and \emph{algorithmic robustness}, which ensures recourse remains valid under various model shifts. 

Since our framework maintains a unified repository of state-of-the-art methods, along with architectural modularity, researchers can inject specialized evaluation modules to benchmark sub-categories across existing methods without modifying the underlying pipelines. To demonstrate this, we instantiate two specialized evaluation suites utilizing the \textbf{German credit}~\citep{german_dataset} and \textbf{South German credit}~\citep{corrected_german_dataset} datasets, the latter specifically for our robustness benchmarking.

\paragraph{Diversity and Set Generation:}
Rather than prescribing a single optimal counterfactual, several approaches propose generating a diverse set of valid options, granting the affected individual the agency to select the path that best aligns with their unobserved personal preferences. Evaluating these methods requires a shift from instance-level distances to set-level distribution metrics. 

In our diversity benchmark, methods generate up to $k=5$ counterfactuals per factual instance $x$, forming a set $C(x)$ where $|C(x)| = |x|*k$. We filter this to a valid subset $V(x)$ where $|V(x)| = m$. To quantify performance, \RecourseBench evaluates:
\begin{itemize}
    \item \textbf{Set Validity:} The fraction of generated counterfactuals that successfully cross the decision boundary, defined as $|V(x)| / |C(x)|$.
    \item \textbf{Unique Fraction:} To ensure methods do not simply return identical duplicates, we compute the number of uniquely distinct valid counterfactuals normalized by $m$.
    \item \textbf{Pairwise Diversity:} Measures the spread of the valid options. We compute the mean pairwise distances (e.g., $\frac{2}{m(m-1)} \sum_{i < j} \|c_i - c_j\|_1$). 
    \item \textbf{Set-to-Factual Proximity:} Measures the mean distance from the valid counterfactuals back to the original factual instance $x$. This distinguishes between sets that are ``diverse but excessively costly'' and those that provide variation while remaining actionable.
\end{itemize}
As shown in \Cref{tab:diversity_summary_combined}, this modular evaluation allows us to compare set-generating methods like DiCE, DiverseDist, COLS, and CEMSP on a unified playing field.

\paragraph{Robustness and Future Validity:}
Algorithmic recourse is inherently temporal; a counterfactual that successfully flips a prediction under a current model ($f_{\text{current}}$) may easily be invalidated if the model is retrained on shifted data ($f_{\text{future}}$). For this benchmark, we utilize the standard German dataset as the current distribution and Southern German as the shifted future distribution, maintaining an identical preprocessing schema across both to ensure spatial consistency. We define \textbf{Future Validity} as the proportion of originally successful counterfactuals that remain valid under the future model:
\begin{equation}
    \text{Future Validity} = \frac{1}{N} \sum_{i=1}^{N} \mathbf{1} \bigl[f_{\text{future}}(c_i) = y_{\text{desired}} \land f_{\text{current}}(c_i) = y_{\text{desired}}\bigr]
\end{equation}
The evaluation of several robustness-aware methods is shown in~\Cref{tab:future_robustness}, on both the linear and MLP model architectures.

\begin{table*}[htbp]
\centering
\caption{Diversity Benchmark Summary (Linear vs. MLP) on the German dataset.}
\label{tab:diversity_summary_combined}
\resizebox{\textwidth}{!}{%
\begin{tabular}{lrrrrrrrrrrrr}
\toprule
& \multicolumn{6}{c}{\textbf{Linear Model}} & \multicolumn{6}{c}{\textbf{MLP Model}} \\
\cmidrule(lr){2-7} \cmidrule(lr){8-13}
Method & Val $\uparrow$ & Set Val. $\uparrow$ & Uniq. Frac. $\uparrow$ & Pair. L2 $\uparrow$ & Set L2 $\downarrow$ & Time (s) $\downarrow$ & Val $\uparrow$ & Set Val. $\uparrow$ & Uniq. Frac. $\uparrow$ & Pair. L2 $\uparrow$ & Set L2 $\downarrow$ & Time (s) $\downarrow$ \\
\midrule
CEMSP & \underline{0.380} & 0.380 & \textbf{1.000} & \textbf{1.246} & \textbf{0.752} & \textbf{6.865} & \underline{0.740} & 0.740 & \textbf{1.000} & \underline{0.919} & 0.928 & \underline{7.447} \\
EMC-COLS & \textbf{1.000} & \underline{0.976} & \textbf{1.000} & \underline{0.868} & \underline{0.811} & \underline{93.650} & \textbf{1.000} & \underline{0.976} & \textbf{1.000} & 0.883 & \underline{0.799} & 118.761 \\
DiCE & \textbf{1.000} & \textbf{1.000} & \textbf{1.000} & 0.794 & 1.023 & 341.167 & \textbf{1.000} & \textbf{1.000} & \textbf{1.000} & \textbf{1.073} & 1.104 & 438.040 \\
Diverse Dist. & - & - & - & - & - & - & \textbf{1.000} & \textbf{1.000} & \textbf{1.000} & 0.263 & \textbf{0.227} & \textbf{1.182} \\
\bottomrule
\end{tabular}%
}
\end{table*}

\begin{table*}[htbp]
\centering
\scriptsize
\caption{Future Robustness Summary (Linear vs. MLP). Current validity is calculated using the German Credit dataset, and future validity is calculated using the South German Credit dataset.}
\label{tab:future_robustness}
\resizebox{0.8\textwidth}{!}{%
\begin{tabular}{lrrrrrr}
    \toprule
    & \multicolumn{3}{c}{\textbf{Linear Model}} & \multicolumn{3}{c}{\textbf{MLP Model}} \\
    \cmidrule(lr){2-4} \cmidrule(lr){5-7}
    Method & Val $\uparrow$ & Future Val $\uparrow$ & Time (s) $\downarrow$ & Val $\uparrow$ & Future Val $\uparrow$ & Time (s) $\downarrow$\\
    \midrule
    APAS & - & - & - & \underline{0.920} & 0.420 & \underline{14.753} \\
    CVAS & \textbf{1.000} & \textbf{1.000} & \underline{20.512} & \underline{0.920} & 0.700 & 18.981 \\
    LARR & \textbf{1.000} & \textbf{1.000} & \textbf{2.323} & \textbf{1.000} & \textbf{1.000} & \textbf{4.995} \\
    PROBE & \textbf{1.000} & 0.760 & 29.278 & 0.600 & 0.340 & 579.783 \\
    RBR & \textbf{1.000} & \underline{0.980} & 151.438 & \textbf{1.000} & \underline{0.760} & 170.075 \\
    ROAR & \textbf{1.000} & \textbf{1.000} & 106.204 & 0.900 & 0.680 & 140.621 \\
    TReX & \textbf{1.000} & \textbf{1.000} & 36.484 & \textbf{1.000} & 0.700 & 45.907 \\
    \bottomrule
\end{tabular}%
}
\end{table*}

%% file: sections/Discussion.tex
\section{Discussion and Future Work}
\label{sec:discussion}

The empirical results presented in Section \ref{sec:evaluation} reveal a fundamental reality of algorithmic recourse: there is no universal ``best'' method. Instead, recourse generation is governed by inherent trade-offs between competing objectives. By examining the exact metrics, distinct trends emerge regarding algorithmic actionability. For instance, sparsity (measured by the $l_0$ norm) dictates the sheer number of features a user must change. In the COMPAS dataset (which contains 7 actionable features), methods optimized for robust recourse, such as PROBE and ROAR, often yield an $l_0$ of 7.0. In practice, this means a user would need to alter every single mutable feature of their profile to achieve a favorable outcome---a task that is cognitively overwhelming and practically difficult. Conversely, methods like Wachter or CFRL achieve an $l_0$ of 1.0, suggesting a single, highly actionable change. However, as the tables demonstrate, this high sparsity often comes at the cost of plausibility (higher yNN) or vulnerability to future distribution shifts.

\paragraph{Model Complexity:}
Beyond the trade-off between sparsity and robustness, our benchmark exposes the extreme sensitivity of many recourse generators to the underlying model's architecture. As observed in Table \ref{tab:compas_combined}, transitioning from a simple Linear model to a Multi-Layer Perceptron (MLP) incurs a severe penalty on validity for several methods. For instance, on the COMPAS dataset, CFVAE drops from a perfect validity of 1.000 on the linear model to 0.380 on the MLP, while LARR catastrophically fails, dropping to 0.000. This stark contrast suggests that many algorithms either implicitly rely on local linearity or struggle to navigate the complex gradient topography of highly non-linear decision boundaries. It highlights the danger of validating proposed methods solely on simple models, emphasizing that architectural robustness is a critical prerequisite for reliable deployment.

\paragraph{The Practicality of Computational Scalability:}
While the literature overwhelmingly prioritizes optimization metrics like $l_2$ and $l_0$, our results demonstrate that computational scalability remains a major bottleneck for practical applications. In a real-world deployment scenario---such as a web portal providing instantaneous feedback on a rejected loan---latency is just as critical as the quality of the recourse. Table \ref{tab:overall_benchmark} reveals that methods prioritizing robust or diversity guarantees, such as PROBE and DiCE, can exhibit severe computational overhead, taking hundreds of seconds to generate a set of counterfactuals on an MLP model. In contrast, optimization-based or gradient-free methods like Wachter and ArgEnsemble consistently execute in under a second. Practitioners must therefore weigh theoretical guarantees against the strict latency requirements of interactive systems, a dimension that \RecourseBench explicitly captures.

\paragraph{Benchmark Stability:}
We acknowledge that single-run evaluations can sometimes mask underlying algorithmic variance. While conducting multi-seed sweeps across roughly 200 configurations poses a prohibitive computational bottleneck, we conducted a focused variance study to establish the statistical stability of our metrics. Specifically, for the COMPAS dataset, we performed factual sampling across four independent random seeds. This targeted variance analysis allows us to characterize the baseline stability of the evaluated recourse methods without imposing unscalable compute requirements on the broader benchmark. As detailed in \Cref{tab:multiseed_stability_compas} within \Cref{sec:further_benchmark}, which reports the mean and standard deviation for validity and $\ell_2$ distance, there is only minor variance across all compatible methods. This confirms that our evaluation pipeline yields statistically reliable and stable results under fixed algorithmic configurations.

\paragraph{The Role of Hyperparameters:}
When interpreting these benchmark results, it is crucial to recognize that the performance of many integrated methods is highly sensitive to their internal hyperparameters. Given the combinatorial explosion, running a full grid search across the hyperparameters of all \numMethodsFull methods was computationally prohibitive. For this benchmark, we defaulted to the exact hyperparameters reported in the original publications or their official codebases. For example, PROBE was evaluated using its paper-reported recourse invalidation rate of 0.35. In cases where parameters dictate a continuous trade-off without a universally ``optimal'' default, we selected structurally balanced values. For instance, LARR utilizes a parameter $\beta \in [0,1]$ that explicitly controls the trade-off between robustness and consistency; we evaluated it at $\beta=0.5$ to provide a balanced baseline. To ensure complete transparency, every specific parameter choice is explicitly defined within the configuration files in our open-source repository. We strongly encourage practitioners to tune these parameters within the framework to observe how algorithmic behavior shifts under different operational constraints.

\paragraph{Human-Centric Evaluation and Future Directions:}
Consequently, the goal of \RecourseBench is not to crown a single state-of-the-art winner, but rather to provide a comprehensive, comparative framework that allows practitioners to answer the question: \textit{``What algorithm is best for my specific use case?''} While our framework standardizes quantitative metrics like $l_0$ and $l_2$, these distance functions are ultimately proxies for true actionability. Fully assessing the real-world utility of a counterfactual often necessitates human-centric evaluation \citep{tominaga2024} to accurately gauge user cognitive load, fairness, and practical feasibility. 

Looking forward, there are several key areas we plan to address to further expand the framework's utility. First, we aim to extend the preprocessing and evaluation layers to support other data modalities, particularly images and text, which require fundamentally different approaches to measuring proximity, sparsity, and plausibility compared to tabular data. Second, we plan to deeply integrate causality-based recourse methods, expanding our core architecture to natively handle Structural Causal Models (SCMs) and causal graphs alongside their specialized evaluations. Finally, our current benchmark primarily evaluates post hoc recourse generation. A vital and growing paradigm involves non-post hoc techniques---methods that inherently train the predictive model to be recourse-friendly by design. Adapting the framework to systematically evaluate these intrinsically interpretable architectures alongside newly proposed algorithms and diverse datasets will ensure \RecourseBench remains a comprehensive, up-to-date tool for the community.

%% file: sections/Conclusion.tex
\section{Conclusion}
\label{sec:conclusion}
In this work, we introduced \RecourseBench, a highly modular, reproducible, and extensible framework designed to standardize the evaluation of algorithmic recourse with reproducibility and transparency at the forefront. By strictly modularizing datasets, target models, recourse methods, and evaluation metrics, we established a rigorous and level playing field for recourse benchmarking. Through our comprehensive evaluation, we highlighted the inherent trade-offs between sparsity, plausibility, robustness, and computational efficiency across diverse algorithms and model architectures. Supported by an accessible Python API and an interactive web dashboard, \RecourseBench provides practitioners and researchers with a dynamic, open-source environment to rapidly prototype, evaluate, and compare recourse generation techniques tailored to their specific real-world deployment contexts.

%% file: sections/Appendix.tex
\section{Technical appendices and supplementary material}



\subsection{Architecture in details}
\label{sec:architecture}

\begin{figure}[htbp] \centering \includegraphics[width=\textwidth]{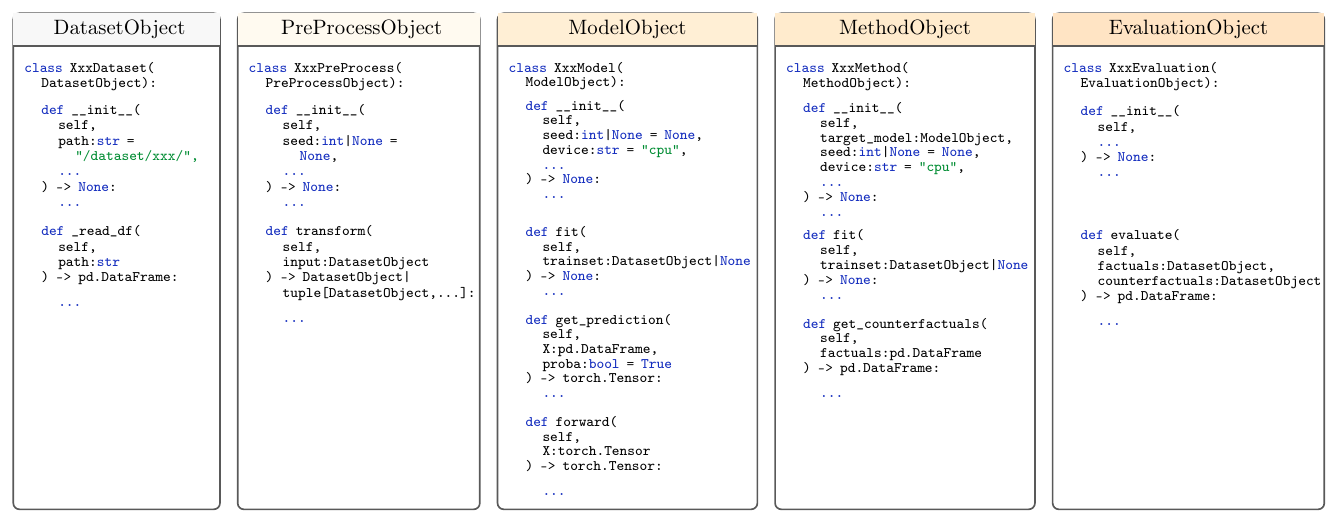} \caption{These are the minimal interfaces to be implemented. Noted that: Settings for components should be implemented inside of their \texttt{\_\_init\_\_()}; datasets' metadata can be dynamically read and loaded from their YAML files; \texttt{ModelObject}'s \texttt{forward()} is used as a gradient-enabled path (\autoref{recourse_methods}). Some functions like \texttt{get\_counterfactuals()} are not using \texttt{DatasetObject} and that's because in these scenarios, \texttt{DatasetObject}s are automatically packed and unpacked by wrapper functions (and the metadata needed here should be read previously in \texttt{fit()}).} \label{fig:code} \end{figure}

The benchmark is driven by a YAML-configured central \texttt{Experiment} workflow. The system heavily relies on Abstract Base Classes (ABCs) to define explicit boundaries.
There're mainly five abstraction layers in which developers can add components: \texttt{Dataset}, \texttt{PreProcess}, \texttt{TargetModel}, \texttt{Method}, \texttt{Evaluation}.
The general dependency data flow is formulated as \autoref{fig:workflow} and code signatures for these five types of components are shown in \autoref{fig:code}.


\textbf{Abstraction Layers \& Interfaces.}
Each component in the five main layers corresponds to a dedicated ABC. Interactions are strictly limited to well-defined public interfaces of the ABCs. Most of the public methods are fixed and only a small part needs to be implemented when introducing new components. By design, the minimal interfaces needed to be implemented are shown in \autoref{fig:code}.
When adding a new component, the number of interfaces that need to be implemented is reduced to as few as one or two, greatly lowering the burden of understanding and implementation while still supporting a complete method training and inference workflow.

\paragraph{Rationale behind the design of \texttt{DatasetObject}:} Compared with frameworks for other tasks such as text or image classification, algorithmic recourse frameworks present a unique challenge: a significant number of recourse papers are working with tabular data, which requires significant amounts of metadata and many different forms of data processing. This scenario is unlike image or text modalities. The former is mainly composed of dense pixels, often with little metadata; the latter is mainly composed of sparse tokens, which generally do not need extra metadata as well. That is the reason we designed DatasetObject as the self-contained message among components.

Except for \texttt{ModelObject} embedded inside \texttt{MethodObject}, \texttt{DatasetObject} is the only thing to be passed between components. Input is a processed DatasetObject, and output is a new DatasetObject. They contain all the metadata needed for the following operations and are automatically packed or unpacked by pre-built functions inside each component. This is beneficial for compatibility checks and keeps each component independent. To prevent undesirable changes to DatasetObject, the \texttt{freeze()} function is introduced, after which no changes can be applied to the object, ensuring re-usability and overall experiment reproducibility.

To avoid other components being altered undesirably, vital internal states are protected. These internal members are copied before returning. All these privilege control features are defined in an abstract class and do not require developers to intervene, while all core methods to be implemented are marked as abstract, enforcing a rigid yet clear development guideline for subclass implementations.

\textbf{Pluggability \& No Hard Coupling.} 
As shown in the interface design, the system operates with minimal assumptions about the underlying algorithms. What's more, components are dynamically plugged into the pipeline using a centralized \texttt{@register} decorator. The \texttt{Experiment} workflow queries this global registry at runtime to instantiate classes defined in the YAML file. Consequently, there are zero hard-coded catalogs, enabling developers to change only one place when adding new components. This greatly reduces omissions and lowers the likelihood of merge conflicts. As a bonus, faulty components can be modified or purged cleanly without affecting anything else.

Compatibility check is achieved by flags inside components. Auxiliary data can all be put into DatasetObject to avoid workflow breach. So there's no hard-coded dependency between two components.

\textbf{Evidence of Scalability} 
The architecture's viability is empirically demonstrated by its ability to easily scale. By adhering to the isolated design and dynamic registry, the framework successfully supports \numMethodsFull recourse methods.
%

\section{Reproducibility Logs}
\label{sec:repro_logs}

Reproducing published results in algorithmic recourse is subject to a compounding set of barriers. The most common ones are proprietary or non-restorable artifacts and unclear documentation. And even in the best scenario that an official implementation exists and hyperparameters are specified, exact numerical reproduction may remain elusive due to sources of non-determinism that are rarely reported in published work. 
Specifically, pseudo-random number sequences are sensitive to execution order; even with an identical seed, changes in the order in which modules are executed can produce divergent trajectories through the random state, leading to materially different outputs. 
Hardware heterogeneity introduces a further confound, as floating-point operator implementations differ across CPUs, GPUs, and other accelerators, yielding numerically distinct results from identical code. 
Finally, software stack versions, particularly deep learning frameworks such as PyTorch, can alter low-level operator behavior across releases, making results obtained under one environment non-transferable to another. 
These factors, individually subtle and collectively compounding, constitute a structural reproducibility challenge. 

\textbf{Limitations on our side.}
Some barriers arise from practical constraints on our side, including commercial software dependencies, prohibitive computational cost due to limited hardware resources, and framework incompatibilities that remain to be addressed in future work.

\subsection{Per-Method Reproduction Notes}
\label{sec:repro_permethod}

This section details the comprehensive claim-based reproduction logs for all \numMethodsFull methods integrated into \RecourseBench. Following the methodology outlined in Section \ref{sec:reproducibility}, each method is assigned a reproducibility taxonomy category based on the availability of original artifacts. Core theoretical and empirical claims were extracted from each original paper and systematically tested within our standardized framework. Greater details about the implementations and their results can be found in our GitHub repository.

\subsection*{APAS}
\noindent\textbf{Taxonomy Categorization:} Reproducible (Code and Data Available) \\
\textbf{Core Claims Evaluated:}
\begin{itemize}
    \item \textit{Qualitative:} APAS verification problem is computationally hard. \textbf{[Unverified (Proof)]}
    \item \textit{Quantitative:} Recourse remains valid under plausible model changes. \textbf{[Verified]}
    \item \textit{Quantitative:} Produces recourse balancing robustness, plausibility, sparsity, and proximity. \textbf{[Verified]}
\end{itemize}
\textbf{Reproduction Summary:} Empirical claims were largely reproduced for the covered scope, with 57 of 58 scalar comparisons passing the 15\% relative-delta threshold. The available CV* diagnostics were strong overall (mean CV* $\approx 0.007$, max CV* $\approx 0.247$). The only failing scalar was Credit $L_0$ distance, reference $0.093$ versus reproduced $0.068$.

\subsection*{Argumentative Ensembling}
\noindent\textbf{Taxonomy Categorization:} Reproducible (Code and Data Available) \\
\textbf{Core Claims Evaluated:}
\begin{itemize}
    \item \textit{Qualitative:} Defines ensemble recourse through coherence/non-emptiness. \textbf{[Unverified (Proof)]}
    \item \textit{Quantitative:} Empirically evaluates accuracy and simplicity across ensemble sizes. \textbf{[Partially Verified]}
\end{itemize}
\textbf{Reproduction Summary:} 170 of 198 scalar accuracy and simplicity comparisons passed the 15\% threshold. CV* diagnostics show moderate agreement overall (mean CV* $\approx 0.060$, max CV* $\approx 0.437$). The main drift occurred in COMPAS simplicity metrics at larger ensemble sizes.

\subsection*{C-CHVAE}
\noindent\textbf{Taxonomy Categorization:} Reproducible (Code and Data Available) \\
\textbf{Core Claims Evaluated:}
\begin{itemize}
    \item \textit{Quantitative:} Generates counterfactuals via VAE-based search. \textbf{[Partially Verified]}
    \item \textit{Quantitative:} Improves realism by searching high-density connected regions. \textbf{[Partially Verified]}
\end{itemize}
\textbf{Reproduction Summary:} The intended C-CHVAE workflow executes and records validity, outlier-rate, connectedness-style, candidate-search, and counterfactual-success metrics. CV* scores and 15\% pass counts are not applicable because the current log does not include precise original scalar targets for strict delta validation.

\subsection*{CEMSP}
\noindent\textbf{Taxonomy Categorization:} Reproducible (Code and Data Available) \\
\textbf{Core Claims Evaluated:}
\begin{itemize}
    \item \textit{Quantitative:} Generates diverse recourse while controlling consistency and sparsity. \textbf{[Partially Verified]}
    \item \textit{Qualitative:} Theoretical support for the count-diversity objective. \textbf{[Unverified (Proof)]}
\end{itemize}
\textbf{Reproduction Summary:} Partially reproduced with 10 of 14 logged scalar comparisons passing the 15\% relative-delta threshold. The associated report contains CV* diagnostics with mean CV* $\approx 0.118$ and max CV* $\approx 0.682$. Discrepancies were observed in mean inconsistency, diversity2, and the mean number of counterfactuals on the HCV/Hepatitis setup.

\subsection*{CFRL}
\noindent\textbf{Taxonomy Categorization:} Robust (Data Available, Independent Implementation) \\
\textbf{Core Claims Evaluated:}
\begin{itemize}
    \item \textit{Quantitative:} Trains a DDPG generator for sparse, in-distribution counterfactuals. \textbf{[Partially Verified]}
    \item \textit{Quantitative:} Supports immutable features and numerical constraints. \textbf{[Partially Verified]}
\end{itemize}
\textbf{Reproduction Summary:} 3 of 7 comparable scalar metrics passed the 15\% threshold. Accuracy, validity, and immutability matched closely, but sparsity and distribution metrics drifted substantially; target-conditional MMD deltas exceeded $0.83$. CV* diagnostics reflect this drift (mean CV* $\approx 0.598$, max CV* $\approx 1.185$).

\subsection*{CFVAE}
\noindent\textbf{Taxonomy Categorization:} Reproducible (Code and Data Available) \\
\textbf{Core Claims Evaluated:}
\begin{itemize}
    \item \textit{Quantitative:} Satisfies causal/actionability constraints more reliably than baselines. \textbf{[Verified]}
    \item \textit{Quantitative:} Preserves proximity while maintaining target-class validity. \textbf{[Verified]}
\end{itemize}
\textbf{Reproduction Summary:} Exhibited exceptionally strong scalar agreement. All 32 scalar comparisons for constraint feasibility and proximity on the Adult dataset passed the 15\% threshold, usually within a 1\% margin. CV* diagnostics were correspondingly low (mean CV* $\approx 0.0012$, max CV* $\approx 0.0069$).

\subsection*{ClaPROAR}
\noindent\textbf{Taxonomy Categorization:} Generalizable (Original in Julia, Incompatible Environment) \\
\textbf{Core Claims Evaluated:}
\begin{itemize}
    \item \textit{Quantitative:} Accounts for how recourse actions affect future model training. \textbf{[Partially Verified]}
    \item \textit{Quantitative:} Reduces model deterioration compared to individual baselines. \textbf{[Partially Verified]}
\end{itemize}
\textbf{Reproduction Summary:} Validated at mechanism level via a scoped synthetic check demonstrating the classifier-preserving penalty. Both variants achieved validity $1.0$, while the ClaPROAR-style variant reduced mean recourse distance from $1.4486$ to $0.8141$. CV* scores and 15\% paper-scalar pass counts are not applicable because the full endogenous retraining and population-dynamics simulations were not reproduced.

\subsection*{CLUE}
\noindent\textbf{Taxonomy Categorization:} Reproducible (Code and Data Available) \\
\textbf{Core Claims Evaluated:}
\begin{itemize}
    \item \textit{Quantitative:} Generates counterfactuals by minimizing predictive uncertainty. \textbf{[Partially Verified]}
    \item \textit{Quantitative:} Reduces epistemic and aleatoric uncertainty. \textbf{[Partially Verified]}
\end{itemize}
\textbf{Reproduction Summary:} 2 of 4 uncertainty scalars passed the 15\% threshold. Epistemic uncertainty reproduced very closely (relative deltas $0.004$ and $0.00015$), while aleatoric uncertainty failed the threshold. CV* diagnostics were mixed (mean CV* $\approx 0.439$, max CV* $\approx 1.591$). Non-tabular image and user-study claims were excluded from programmatic evaluation.

\subsection*{CoGS}
\noindent\textbf{Taxonomy Categorization:} Reproducible (Code and Data Available) \\
\textbf{Core Claims Evaluated:}
\begin{itemize}
    \item \textit{Quantitative:} Quantifies and improves accidental robustness of counterfactuals. \textbf{[Partially Verified]}
    \item \textit{Quantitative:} Evaluates runtime/cost trade-offs across datasets. \textbf{[Unverified]}
\end{itemize}
\textbf{Reproduction Summary:} Robustness-style measurements execute, but the logged run is heavily scoped to Adult, random forest, and very few factuals due to heavy runtimes. CV* scores and 15\% pass counts are not applicable due to being heavily scoped as of now.

\subsection*{EMC-COLS}
\noindent\textbf{Taxonomy Categorization:} Reproducible (Code and Data Available) \\
\textbf{Core Claims Evaluated:}
\begin{itemize}
    \item \textit{Quantitative:} Generates recourse sets improving expected user satisfaction. \textbf{[Partially Verified]}
    \item \textit{Quantitative:} Improves coverage, diversity, proximity, and sparsity. \textbf{[Partially Verified]}
\end{itemize}
\textbf{Reproduction Summary:} 59 of 70 scalar comparisons passed the 15\% threshold. CV* diagnostics show generally moderate agreement with notable outliers (mean CV* $\approx 0.114$, max CV* $\approx 1.591$). Important discrepancies remain in LS Diversity FS@1 and PAC metrics.

\subsection*{CRUDS}
\noindent\textbf{Taxonomy Categorization:} Generalizable (No Original Code or Data Available) \\
\textbf{Core Claims Evaluated:}
\begin{itemize}
    \item \textit{Quantitative:} Gradient-optimization can fail via topology traps/low-density outputs. \textbf{[Partially Verified]}
    \item \textit{Quantitative:} Target-class sampling improves constraint satisfaction. \textbf{[Partially Verified]}
\end{itemize}
\textbf{Reproduction Summary:} A scoped synthetic mixture check validated the core conceptual claim: boundary-distance recourse landed in a low-density region while target-support sampling avoided this failure mode. CV* scores and 15\% paper-scalar pass counts are not applicable because original tables were not replicated.

\subsection*{CVAS}
\noindent\textbf{Taxonomy Categorization:} Reproducible (Code and Data Available) \\
\textbf{Core Claims Evaluated:}
\begin{itemize}
    \item \textit{Quantitative:} Generates feasible recourses with strong current/future validity. \textbf{[Partially Verified]}
    \item \textit{Quantitative:} Evaluates cost and feasibility across multiple tabular datasets. \textbf{[Partially Verified]}
\end{itemize}
\textbf{Reproduction Summary:} Strong structural reproduction with 13 of 18 scalar comparisons passing the 15\% threshold. Mean targets for future-validity and cost aligned very closely, including German future-validity mean and SBA cost mean. CV* diagnostics were affected by divergent standard-deviation metrics (mean CV* $\approx 0.273$, max CV* $\approx 1.591$).

\subsection*{DiCE}
\noindent\textbf{Taxonomy Categorization:} Reproducible (Code and Data Available) \\
\textbf{Core Claims Evaluated:}
\begin{itemize}
    \item \textit{Quantitative:} Generates diverse counterfactuals maintaining high validity. \textbf{[Partially Verified]}
    \item \textit{Quantitative:} Approximates local decision boundaries for feasibility. \textbf{[Partially Verified]}
\end{itemize}
\textbf{Reproduction Summary:} Validated directionally within a scoped COMPAS setting, demonstrating that DiCE yields higher diversity than no-diversity/random-initialization variants while maintaining high validity ($0.965$--$1.0$). CV* scores and 15\% pass counts are not applicable because no current scalar-delta report with paper targets is available.

\subsection*{DiverseDist}
\noindent\textbf{Taxonomy Categorization:} Reproducible (Code and Data Available) \\
\textbf{Core Claims Evaluated:}
\begin{itemize}
    \item \textit{Quantitative:} Generates valid diverse counterfactual sets across norms. \textbf{[Partially Verified]}
    \item \textit{Quantitative:} Improves $k$-distance and set-distance over baselines. \textbf{[Partially Verified]}
\end{itemize}
\textbf{Reproduction Summary:} Algorithmic validity was reliably reproduced, but only 57 of 132 scalar comparisons passed the 15\% threshold. Distance and timing values drifted significantly depending on optimization settings. CV* diagnostics were correspondingly mixed (mean CV* $\approx 0.272$, max CV* $\approx 1.490$).

\subsection*{FACE}
\noindent\textbf{Taxonomy Categorization:} Generalizable (No Original Code or Data Available) \\
\textbf{Core Claims Evaluated:}
\begin{itemize}
    \item \textit{Quantitative:} Finds feasible counterfactuals via data-manifold graphs. \textbf{[Partially Verified]}
    \item \textit{Quantitative:} Prefers realistic counterfactuals using density-aware construction. \textbf{[Partially Verified]}
\end{itemize}
\textbf{Reproduction Summary:} The conceptual mechanism was successfully validated on a synthetic setup, confirming graph-path behavior and reachability. CV* scores and 15\% pass counts are not applicable because exact original scalar targets are unavailable.

\subsection*{FeatureTweak}
\noindent\textbf{Taxonomy Categorization:} Generalizable (Proprietary Data Unavailable) \\
\textbf{Core Claims Evaluated:}
\begin{itemize}
    \item \textit{Quantitative:} Identifies actionable changes altering tree-ensemble outcomes. \textbf{[Partially Verified]}
    \item \textit{Qualitative:} Distinguishes adjustable from static features. \textbf{[Partially Verified]}
\end{itemize}
\textbf{Reproduction Summary:} Due to proprietary data, exact replication was impossible. A synthetic tree-ensemble check validated the mechanism, flipping all three scoped negative factuals while producing zero static-feature violations. CV* scores are not reported; the only scoped scalar with a target passed, but no original paper scalar table was reproduced.

\subsection*{Gravitational}
\noindent\textbf{Taxonomy Categorization:} Generalizable (Original in Julia, Incompatible Environment) \\
\textbf{Core Claims Evaluated:}
\begin{itemize}
    \item \textit{Quantitative:} Reduces downstream model deterioration relative to baselines. \textbf{[Partially Verified]}
    \item \textit{Quantitative:} Models individual recourse inducing distribution shift. \textbf{[Unverified]}
\end{itemize}
\textbf{Reproduction Summary:} The local target-domain penalty mechanism was verified: validity remained $1.0$ and mean target-center distance dropped from $2.0529$ to $0.0011$. CV* scores and 15\% paper-scalar pass counts are not applicable because downstream model-shift and repeated-retraining simulations were omitted.

\subsection*{Growing Spheres (GS)}
\noindent\textbf{Taxonomy Categorization:} Reproducible (Code and Data Available) \\
\textbf{Core Claims Evaluated:}
\begin{itemize}
    \item \textit{Quantitative:} Generates recourse by expanding a search sphere. \textbf{[Partially Verified]}
    \item \textit{Quantitative:} Finds sparse feature changes flipping classifier decisions. \textbf{[Partially Verified]}
\end{itemize}
\textbf{Reproduction Summary:} 1 of 4 comparable scalars passed the 15\% threshold. Predictive AUC matched closely (relative delta $0.008$; CV* $\approx 0.006$), but $L_0$ sparsity metrics diverged, including max $L_0$ with relative delta $0.685$. Overall CV* diagnostics were mixed (mean CV* $\approx 0.301$, max CV* $\approx 0.829$).

\subsection*{LARR}
\noindent\textbf{Taxonomy Categorization:} Reproducible (Code and Data Available) \\
\textbf{Core Claims Evaluated:}
\begin{itemize}
    \item \textit{Quantitative:} Recourse follows the intended robustness-consistency tradeoff. \textbf{[Partially Verified]}
    \item \textit{Quantitative:} Evaluates smoothness, prediction-error, and validity costs. \textbf{[Partially Verified]}
\end{itemize}
\textbf{Reproduction Summary:} Tradeoff endpoints behaved exactly as expected across all six dataset/model settings: maximum $\beta$ yielded zero average robustness and minimum $\beta$ yielded zero average consistency. CV* scores and 15\% pass counts are not applicable because the current LARR report validates sweep behavior rather than comparing against explicit scalar paper targets. Linear-model validity was strong, while MLP variants showed mixed validity-cost behavior.

\subsection*{PROBE}
\noindent\textbf{Taxonomy Categorization:} Reproducible (Code and Data Available) \\
\textbf{Core Claims Evaluated:}
\begin{itemize}
    \item \textit{Quantitative:} Accounts for uncertainty in future model behavior. \textbf{[Partially Verified]}
    \item \textit{Quantitative:} Improves reliability of recourse under model variation. \textbf{[Partially Verified]}
\end{itemize}
\textbf{Reproduction Summary:} 26 of 36 comparisons passed the 15\% threshold. Means align well for some linear models, including Adult linear AIR mean with relative delta $0.025$, but notable discrepancies exist in MLP costs and standard deviations. CV* diagnostics were moderate with outliers (mean CV* $\approx 0.144$, max CV* $\approx 1.084$).

\subsection*{ProPlace}
\noindent\textbf{Taxonomy Categorization:} Reproducible (Code and Data Available) \\
\textbf{Core Claims Evaluated:}
\begin{itemize}
    \item \textit{Quantitative:} Generates plausible robust recourse under model multiplicity. \textbf{[Verified]}
    \item \textit{Quantitative:} Preserves validity, $L_1$ cost, and LOF across datasets. \textbf{[Verified]}
\end{itemize}
\textbf{Reproduction Summary:} Demonstrated strong empirical reproduction, with 18 of 20 scalar comparisons passing the 15\% threshold. CV* diagnostics were low overall (mean CV* $\approx 0.040$, max CV* $\approx 0.328$). Minor deviations were isolated to Adult and HELOC $L_1$ costs.

\subsection*{RBR}
\noindent\textbf{Taxonomy Categorization:} Reproducible (Code and Data Available) \\
\textbf{Core Claims Evaluated:}
\begin{itemize}
    \item \textit{Quantitative:} Formulates recourse robust to shifts between current and future classifiers. \textbf{[Partially Verified]}
    \item \textit{Quantitative:} Evaluates validity and accuracy across robustness sweeps. \textbf{[Partially Verified]}
\end{itemize}
\textbf{Reproduction Summary:} CV* validation is not possible since the original report lacks scalar target values. The presented figures were reproduced and display similar trends when performing $\epsilon$ sweeps.

\subsection*{REVISE}
\noindent\textbf{Taxonomy Categorization:} Generalizable (No Original Code or Data Available) \\
\textbf{Core Claims Evaluated:}
\begin{itemize}
    \item \textit{Quantitative:} Generates realistic recourse by optimizing in a generative latent space. \textbf{[Partially Verified]}
    \item \textit{Quantitative:} Compares latent-manifold recourse against input-space optimization. \textbf{[Partially Verified]}
\end{itemize}
\textbf{Reproduction Summary:} Conceptually verified via a scoped manifold check. Latent-space optimization produced valid recourse with target probability $0.9925$ and zero manifold residual, while direct input-space optimization had manifold residual $0.3443$. CV* scores are not reported; the scoped zero-residual scalar passed, but original paper scalar tables were not reproduced.

\subsection*{ROAR}
\noindent\textbf{Taxonomy Categorization:} Reproducible (Code and Data Available) \\
\textbf{Core Claims Evaluated:}
\begin{itemize}
    \item \textit{Quantitative:} Generates recourse valid under plausible model shifts. \textbf{[Partially Verified]}
    \item \textit{Quantitative:} Evaluates future validity, cost, and variation metrics. \textbf{[Partially Verified]}
\end{itemize}
\textbf{Reproduction Summary:} 36 of 72 metrics passed the 15\% threshold. Validity means generally aligned closely, but the optimization cost (PFC) metric frequently drifted. CV* diagnostics show substantial outliers (mean CV* $\approx 0.378$, max CV* $\approx 1.591$).

\subsection*{SNS}
\noindent\textbf{Taxonomy Categorization:} Robust (Data Available, Independent Experiment Implementation) \\
\textbf{Core Claims Evaluated:}
\begin{itemize}
    \item \textit{Quantitative:} Stable-neighbor recourse reduces invalidation under model changes. \textbf{[Partially Verified]}
    \item \textit{Quantitative:} Balances invalidation against recourse cost. \textbf{[Partially Verified]}
\end{itemize}
\textbf{Reproduction Summary:} 17 of 36 metrics passed the 15\% threshold. Directional invalidation trends support the claims, including German Credit SNS leave-one-out invalidation matching $0.0$, but zero-reference mismatches caused relative-delta failures. CV* diagnostics reflect these failures (mean CV* $\approx 0.542$, max CV* $\approx 1.591$).

\subsection*{TReX}
\noindent\textbf{Taxonomy Categorization:} Reproducible (Code and Data Available) \\
\textbf{Core Claims Evaluated:}
\begin{itemize}
    \item \textit{Quantitative:} Improves future validity while controlling cost and likeness/LOF. \textbf{[Partially Verified]}
    \item \textit{Qualitative:} Theoretical support for future-valid recourse. \textbf{[Unverified (Proof)]}
\end{itemize}
\textbf{Reproduction Summary:} 16 of 24 metrics passed the 15\% threshold. Leave-one-out future validity matched exceptionally well, including HELOC L2 leave-one-out validity with relative delta $0.009$. LOF metrics exhibited the largest drift, yielding high CV* outliers (mean CV* $\approx 0.912$, max CV* $\approx 10.231$).

\subsection*{Wachter}
\noindent\textbf{Taxonomy Categorization:} Generalizable (No Original Code or Data Available) \\
\textbf{Core Claims Evaluated:}
\begin{itemize}
    \item \textit{Quantitative:} Generates closest possible worlds that change predictions. \textbf{[Partially Verified]}
    \item \textit{Quantitative:} Balances prediction loss and distance for interpretability. \textbf{[Partially Verified]}
\end{itemize}
\textbf{Reproduction Summary:} A deterministic scoped optimizer check conceptually validated the claim. Both unweighted and MAD-weighted variants crossed the decision threshold, with target probabilities $0.5553$ and $0.5457$. The MAD-weighted $L_1$ variant reduced movement in the high-cost second feature from $0.2383$ to $0.0034$. CV* scores and 15\% paper-scalar pass counts are not applicable because the original LSAT/Pima examples and scalar tables were not reproduced.

\section{Further Benchmarking Analysis}
\label{sec:further_benchmark}

Benchmarks were executed in CPU mode on a laptop with a 13th Gen Intel Core i5-13450HX CPU, 10 physical cores, 16 logical processors, and 12 GB RAM. The laptop also has access to an NVIDIA GeForce RTX 4050 Laptop GPU.

\begin{table}[htbp]
\centering
\caption{Target Model Predictive Performance. Baseline accuracy and AUC scores for the four core datasets across all evaluated architectures.}
\label{tab:model_performance}
\begin{tabular}{llrr}
\toprule
\textbf{Dataset} & \textbf{Model Architecture} & \textbf{Accuracy} & \textbf{AUC} \\
\midrule
\multirow{4}{*}{Adult} & Linear & 0.819 & 0.903 \\
 & MLP & 0.824 & 0.898 \\
 & Random Forest & 0.815 & 0.895 \\
 & Bayesian MLP & 0.824 & 0.908 \\
\midrule
\multirow{4}{*}{Boston Housing} & Linear & 0.843 & 0.929 \\
 & MLP & 0.863 & 0.937 \\
 & Random Forest & 0.863 & 0.933 \\
 & Bayesian MLP & 0.725 & 0.898 \\
\midrule
\multirow{4}{*}{COMPAS} & Linear & 0.669 & 0.741 \\
 & MLP & 0.694 & 0.742 \\
 & Random Forest & 0.661 & 0.720 \\
 & Bayesian MLP & 0.689 & 0.748 \\
\midrule
\multirow{4}{*}{Credit} & Linear & 0.710 & 0.766 \\
 & MLP & 0.714 & 0.777 \\
 & Random Forest & 0.702 & 0.763 \\
 & Bayesian MLP & 0.722 & 0.779 \\
\bottomrule
\end{tabular}
\end{table}

\begin{table*}[htbp]
\scriptsize
\centering
\caption{Benchmarking Results: Adult Dataset (Linear vs. MLP).}
\label{tab:adult_combined}
\resizebox{\textwidth}{!}{%
\begin{tabular}{lrrrrrrrrrr}
\toprule
& \multicolumn{5}{c}{\textbf{Linear Model}} & \multicolumn{5}{c}{\textbf{MLP Model}} \\
\cmidrule(lr){2-6} \cmidrule(lr){7-11}
Method & Val $\uparrow$ & L0 $\downarrow$ & L2 $\downarrow$ & yNN-5 $\downarrow$ & Time (s) $\downarrow$ & Val $\uparrow$ & L0 $\downarrow$ & L2 $\downarrow$ & yNN-5 $\downarrow$ & Time (s) $\downarrow$ \\
\midrule
APAS & - & - & - & - & - & \textbf{1.000} & 3.520 & \textbf{0.074} & 0.389 & 9.447 \\
ArgEnsembling & \textbf{1.000} & 3.840 & 0.902 & \underline{0.318} & 1.071 & \textbf{1.000} & 3.780 & 0.939 & 0.277 & 1.090 \\
C-CHVAE & \textbf{1.000} & 6.140 & 1.834 & 1.691 & 0.895 & \textbf{1.000} & 6.480 & 1.964 & 1.817 & 0.902 \\
CEMSP & \underline{0.980} & 2.633 & 1.453 & 1.349 & 1.520 & 0.860 & 2.488 & 1.508 & 1.478 & 1.007 \\
CFRL & \textbf{1.000} & 10.180 & 2.402 & \textbf{0.136} & \underline{0.873} & \textbf{1.000} & 11.000 & 2.612 & \textbf{0.148} & 0.815 \\
CFVAE & \textbf{1.000} & 10.440 & 2.742 & 1.471 & 0.879 & \textbf{1.000} & 12.840 & 3.268 & 1.631 & \underline{0.805} \\
ClaPROAR & \textbf{1.000} & 51.000 & \underline{0.164} & 0.406 & 0.966 & 0.680 & 51.000 & \underline{0.086} & 0.356 & 0.832 \\
CoGS & \textbf{1.000} & \underline{1.120} & 0.437 & 0.613 & 1.752 & \textbf{1.000} & \textbf{1.000} & 0.425 & 0.633 & 1.791 \\
EMC-COLS & \textbf{1.000} & 1.760 & 0.510 & 0.565 & 108.676 & \textbf{1.000} & \underline{1.840} & 0.620 & 0.869 & 109.260 \\
CRUDS & 0.260 & 8.692 & 2.141 & 0.383 & 4.143 & 0.340 & 10.059 & 2.435 & 0.302 & 4.616 \\
CVAS & 0.960 & \textbf{1.000} & 0.216 & 0.432 & 2.642 & \underline{0.980} & \textbf{1.000} & 0.237 & 0.473 & 2.308 \\
DiCE & \textbf{1.000} & 5.060 & 0.324 & 0.480 & 255.737 & \textbf{1.000} & 9.960 & 2.164 & 1.420 & 264.917 \\
Diverse Dist. & - & - & - & - & - & \textbf{1.000} & 3.880 & 0.515 & 0.434 & \textbf{0.789} \\
FACE & - & - & - & - & - & - & - & - & - & - \\
Gravitational & \textbf{1.000} & 10.720 & 2.582 & 0.439 & 0.928 & \textbf{1.000} & 11.180 & 2.672 & 0.602 & 0.898 \\
GS & \textbf{1.000} & 16.620 & 3.960 & 3.400 & 2.323 & \textbf{1.000} & 16.880 & 4.037 & 3.478 & 5.841 \\
LARR & \textbf{1.000} & 42.480 & 0.296 & 0.501 & 1.297 & 0.720 & 28.444 & 1.957 & 1.930 & 6.684 \\
PROBE & \textbf{1.000} & 51.000 & 0.268 & 0.485 & 28.996 & 0.260 & 29.769 & 0.107 & \underline{0.261} & 676.780 \\
Proplace & \textbf{1.000} & 3.720 & 0.894 & 0.300 & 14.467 & \textbf{1.000} & 4.400 & 1.076 & 0.361 & 238.508 \\
RBR & \textbf{1.000} & 51.000 & 0.238 & 0.444 & 133.400 & \textbf{1.000} & 51.000 & 0.248 & 0.477 & 204.461 \\
Revise & 0.800 & 5.125 & 0.796 & 0.407 & 4.024 & 0.360 & 5.167 & 0.852 & 0.453 & 4.249 \\
ROAR & \textbf{1.000} & 51.000 & 0.239 & 0.449 & 41.075 & 0.960 & 51.000 & 0.775 & 0.895 & 49.138 \\
SNS & - & - & - & - & - & \textbf{1.000} & 42.340 & 0.174 & 0.443 & 147.893 \\
TReX & \textbf{1.000} & 22.960 & 0.199 & 0.428 & 17.646 & \textbf{1.000} & 33.720 & 0.200 & 0.457 & 23.091 \\
Wachter & 0.680 & 4.000 & \textbf{0.084} & 0.319 & \textbf{0.846} & 0.440 & 3.773 & 0.109 & 0.406 & 7.965 \\
\bottomrule
\end{tabular}%
}
\end{table*}

\begin{table*}[htbp]
\scriptsize
\centering
\caption{Benchmarking Results: Boston Housing Dataset (Linear vs. MLP)}
\label{tab:boston_combined}
\resizebox{\textwidth}{!}{%
\begin{tabular}{lrrrrrrrrrr}
\toprule
& \multicolumn{5}{c}{\textbf{Linear Model}} & \multicolumn{5}{c}{\textbf{MLP Model}} \\
\cmidrule(lr){2-6} \cmidrule(lr){7-11}
Method & Val $\uparrow$ & L0 $\downarrow$ & L2 $\downarrow$ & yNN-5 $\downarrow$ & Time (s) $\downarrow$ & Val $\uparrow$ & L0 $\downarrow$ & L2 $\downarrow$ & yNN-5 $\downarrow$ & Time (s) $\downarrow$ \\
\midrule
APAS & - & - & - & - & - & \textbf{1.000} & 9.500 & \underline{0.187} & 0.264 & 41.189 \\
ArgEnsembling & \textbf{1.000} & 7.180 & 0.486 & \textbf{0.212} & 0.869 & \textbf{1.000} & 6.848 & 0.390 & \textbf{0.177} & 0.820 \\
C-CHVAE & \textbf{1.000} & 11.800 & 1.355 & 1.028 & 0.704 & \textbf{1.000} & 11.413 & \textbf{1.143} & 0.883 & 0.662 \\
CEMSP & \textbf{1.000} & 1.760 & 0.452 & \underline{0.304} & 1.043 & \textbf{1.000} & 1.435 & 0.349 & 0.295 & 0.979 \\
CFRL & \textbf{1.000} & 10.960 & 0.883 & 0.387 & \underline{0.607} & \textbf{1.000} & 10.978 & 0.737 & 0.376 & 0.564 \\
CFVAE & 0.860 & 12.419 & 1.316 & 0.487 & \textbf{0.589} & 0.761 & 12.400 & 1.331 & 0.480 & \textbf{0.524} \\
ClaPROAR & \underline{0.980} & 21.000 & 0.397 & 0.389 & 0.700 & 0.935 & 21.000 & 0.196 & 0.272 & 0.725 \\
CoGS & \textbf{1.000} & \underline{1.580} & 0.621 & 0.614 & 1.394 & \textbf{1.000} & \underline{1.326} & 0.413 & 0.381 & 1.315 \\
EMC-COLS & \textbf{1.000} & 3.760 & 0.495 & 0.429 & 103.721 & \textbf{1.000} & 2.152 & 0.287 & 0.311 & 95.732 \\
CRUDS & \textbf{1.000} & 12.300 & 2.121 & 1.666 & 3.575 & \textbf{1.000} & 12.870 & 2.195 & 1.310 & 3.732 \\
CVAS & \textbf{1.000} & \textbf{1.000} & 0.495 & 0.464 & 1.670 & \underline{0.978} & \textbf{1.000} & 0.365 & 0.359 & 1.542 \\
DiCE & \textbf{1.000} & 11.220 & 1.203 & 0.751 & 250.485 & \textbf{1.000} & 11.369 & 0.931 & 0.541 & 241.457 \\
Diverse Dist. & - & - & - & - & - & \textbf{1.000} & 7.239 & 0.292 & \underline{0.213} & \underline{0.561} \\
Gravitational & \textbf{1.000} & 12.780 & 1.572 & 0.333 & 0.692 & \textbf{1.000} & 12.783 & 1.588 & 0.332 & 0.689 \\
GS & \textbf{1.000} & 3.400 & 1.690 & 1.607 & 5.906 & \textbf{1.000} & 3.022 & 1.636 & 1.518 & 5.861 \\
LARR & \textbf{1.000} & 10.860 & 0.991 & 0.870 & 0.872 & \textbf{1.000} & 14.196 & 1.790 & 1.448 & 2.377 \\
PROBE & \textbf{1.000} & 21.000 & \underline{0.328} & 0.410 & 19.509 & 0.804 & 18.568 & 0.254 & 0.341 & 312.705 \\
Proplace & \textbf{1.000} & 7.580 & 0.351 & 0.235 & 4.542 & \textbf{1.000} & 9.196 & 1.030 & 0.249 & 2203.524 \\
RBR & \textbf{1.000} & 21.000 & 0.424 & 0.346 & 147.059 & \textbf{1.000} & 21.000 & 0.337 & 0.269 & 136.267 \\
Revise & \textbf{1.000} & 12.140 & 1.293 & 0.681 & 3.305 & \textbf{1.000} & 12.087 & 1.241 & 0.581 & 3.590 \\
ROAR & \textbf{1.000} & 21.000 & 0.580 & 0.561 & 24.608 & \textbf{1.000} & 21.000 & 1.015 & 0.899 & 48.250 \\
SNS & - & - & - & - & - & \textbf{1.000} & 17.978 & 0.253 & 0.289 & 134.398 \\
TReX & \textbf{1.000} & 17.100 & 0.399 & 0.377 & 17.692 & \textbf{1.000} & 17.587 & 0.283 & 0.301 & 20.883 \\
Wachter & \underline{0.980} & 9.837 & \textbf{0.284} & 0.347 & 0.822 & 0.957 & 9.977 & 0.240 & 0.305 & 0.810 \\
\bottomrule
\end{tabular}%
}
\end{table*}

\begin{table}[htbp]
\centering
\scriptsize
\caption{Benchmark Results: Feature Tweak on Random Forest models.}
\label{tab:feature_tweak_rf}
\resizebox{\columnwidth}{!}{%
\begin{tabular}{llrrrrr}
\toprule
Dataset & Model & Val $\uparrow$ & L0 $\downarrow$ & L2 $\downarrow$ & yNN-5 $\downarrow$ & Runtime (s) $\downarrow$ \\
\midrule
Adult & randomforest & 1.000 & 1.180 & 0.077 & 0.649 & 32302.079 \\
Boston Housing & randomforest & 1.000 & 3.560 & 0.209 & 0.268 & 905.139 \\
COMPAS & randomforest & 1.000 & 1.000 & 0.074 & 0.010 & 4851.984 \\
Credit & randomforest & 1.000 & 2.820 & 0.114 & 0.153 & 40757.160 \\
\bottomrule
\end{tabular}%
}
\end{table}

\begin{table}[htbp]
\centering
\scriptsize
\caption{Benchmark Results: CLUE on Bayesian MLP models.}
\label{tab:clue_mlp_bayesian}
\resizebox{\columnwidth}{!}{%
\begin{tabular}{llrrrrr}
\toprule
Dataset & Model & Val $\uparrow$ & L0 $\downarrow$ & L2 $\downarrow$ & yNN-5 $\downarrow$ & Runtime (s) $\downarrow$ \\
\midrule
Adult & mlp\_bayesian & 0.500 & 8.680 & 1.178 & 0.229 & 2.255 \\
Boston Housing & mlp\_bayesian & 0.380 & 13.684 & 2.096 & 1.473 & 2.155 \\
COMPAS & mlp\_bayesian & 0.340 & 4.471 & 0.881 & 0.568 & 1.535 \\
Credit & mlp\_bayesian & 0.100 & 12.400 & 1.219 & 0.758 & 1.602 \\
\bottomrule
\end{tabular}%
}
\end{table}

\begin{table*}[htbp]
\centering
\caption{Multiseed Stability Summary (Linear vs. MLP) on COMPAS. Missing values (-) indicate method incompatibility or instances where the model failed to produce valid counterfactuals.}
\label{tab:multiseed_stability_compas}
\resizebox{\textwidth}{!}{%
\begin{tabular}{lrrrrrrrr}
\toprule
& \multicolumn{4}{c}{\textbf{Linear Model}} & \multicolumn{4}{c}{\textbf{MLP Model}} \\
\cmidrule(lr){2-5} \cmidrule(lr){6-9}
Method & Val Mean & Val Std & L2 Mean & L2 Std & Val Mean & Val Std & L2 Mean & L2 Std \\
\midrule
APAS & - & - & - & - & \textbf{1.000} & 0.000 & 0.191 & 0.047 \\
ArgEnsembling & \textbf{1.000} & 0.000 & 0.195 & 0.019 & \textbf{1.000} & 0.000 & 0.202 & 0.047 \\
C-CHVAE & \textbf{1.000} & 0.000 & 0.723 & 0.142 & \textbf{1.000} & 0.000 & 0.576 & 0.342 \\
CEMSP & 0.665 & 0.065 & 0.177 & 0.005 & 0.625 & 0.090 & 0.172 & 0.031 \\
CFRL & \textbf{1.000} & 0.000 & 0.318 & 0.014 & 0.845 & 0.046 & 0.171 & 0.021 \\
CFVAE & \textbf{1.000} & 0.000 & 1.609 & 0.071 & 0.770 & 0.033 & 1.635 & 0.028 \\
ClaPROAR & \textbf{1.000} & 0.000 & 0.445 & 0.008 & 0.820 & 0.032 & 0.195 & 0.009 \\
CoGS & \textbf{1.000} & 0.000 & 0.164 & 0.011 & \textbf{1.000} & 0.000 & 0.194 & 0.047 \\
EMC-COLS & \textbf{1.000} & 0.000 & 0.187 & 0.009 & \textbf{1.000} & 0.000 & 0.227 & 0.047 \\
CRUDS & 0.395 & 0.030 & 0.490 & 0.049 & 0.650 & 0.092 & 0.613 & 0.060 \\
CVAS & \textbf{1.000} & 0.000 & 0.295 & 0.009 & 0.865 & 0.050 & 0.226 & 0.014 \\
DiCE & \textbf{1.000} & 0.000 & 1.245 & 0.079 & \textbf{1.000} & 0.000 & 1.331 & 0.028 \\
Diverse Dist. & - & - & - & - & \textbf{1.000} & 0.000 & 0.181 & 0.036 \\
FACE & 0.975 & 0.017 & 0.208 & 0.012 & 0.910 & 0.030 & 0.123 & 0.024 \\
Gravitational & 0.150 & 0.066 & 0.294 & 0.134 & 0.805 & 0.046 & 0.480 & 0.036 \\
GS & \textbf{1.000} & 0.000 & 0.504 & 0.010 & \textbf{1.000} & 0.000 & 0.470 & 0.028 \\
LARR & \textbf{1.000} & 0.000 & 0.752 & 0.009 & 0.000 & 0.000 & - & - \\
PROBE & \textbf{1.000} & 0.000 & 0.373 & 0.016 & 0.735 & 0.050 & 0.134 & 0.015 \\
Proplace & \textbf{1.000} & 0.000 & 0.177 & 0.008 & \textbf{1.000} & 0.000 & 0.251 & 0.057 \\
RBR & \textbf{1.000} & 0.000 & 0.315 & 0.008 & \textbf{1.000} & 0.000 & 0.271 & 0.015 \\
Revise & \textbf{1.000} & 0.000 & 0.557 & 0.009 & 0.815 & 0.074 & 0.380 & 0.030 \\
ROAR & \textbf{1.000} & 0.000 & 0.660 & 0.011 & 0.995 & 0.009 & 0.429 & 0.023 \\
SNS & - & - & - & - & 0.930 & 0.033 & 0.231 & 0.013 \\
TReX & \textbf{1.000} & 0.000 & 0.432 & 0.009 & 0.965 & 0.022 & 0.444 & 0.026 \\
Wachter & \textbf{1.000} & 0.000 & 0.162 & 0.009 & 0.875 & 0.038 & 0.119 & 0.024 \\
\bottomrule
\end{tabular}%
}
\end{table*}

\clearpage

\section{Benchmark Interface}
\label{sec:interface}

\begin{figure*}[ht]
    \centering
    \begin{subfigure}[t]{\linewidth}
        \centering
        \includegraphics[width=0.9\linewidth]{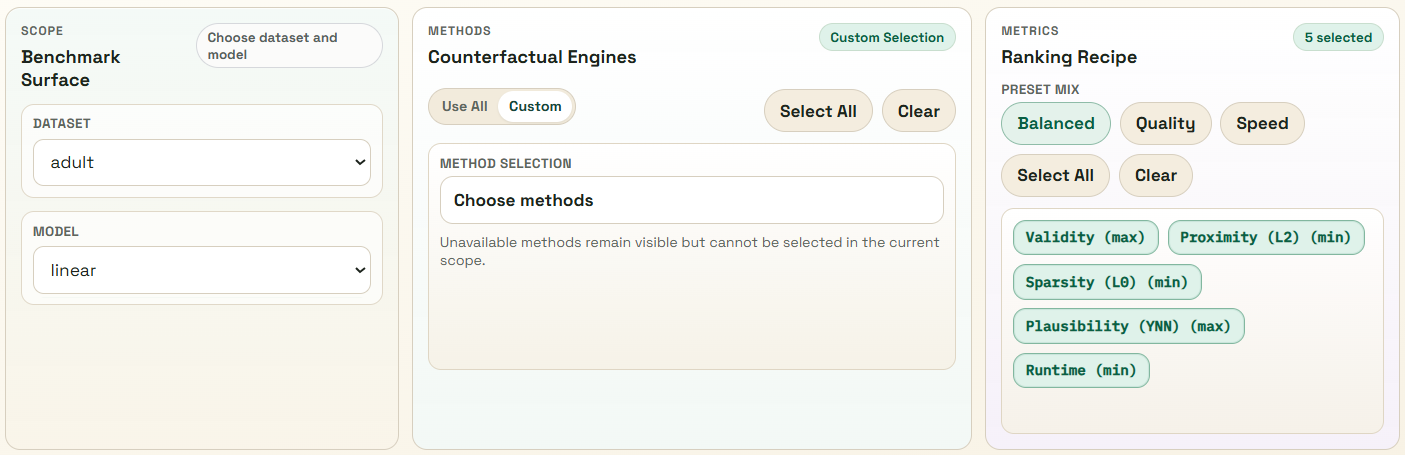}
        \caption{Configuration panel: dataset, model architecture, recourse 
        methods, and evaluation metrics are selected jointly. Incompatible 
        method--model combinations are automatically dimmed.}
        \label{fig:ui_config}
    \end{subfigure}

    \vspace{0.4em}

    \begin{subfigure}[t]{\linewidth}
        \centering
        \includegraphics[width=0.9\linewidth]{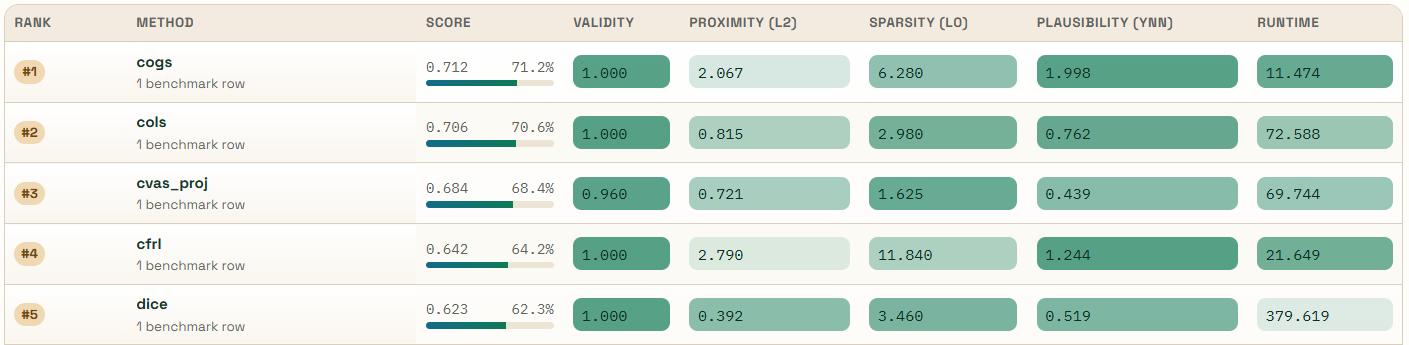}
        \caption{Ranked leaderboard: methods are sorted by aggregate score; 
        per-metric normalized values are shown with color-coded intensity.}
        \label{fig:ui_table}
    \end{subfigure}

    \caption{The \RecourseBench interface. (A) The configuration panel defines the experimental scope; incompatible options are dimmed automatically based on per-method compatibility declarations. (B) The ranked leaderboard presents methods sorted by aggregate score, with per-metric values shown alongside.}
    \label{fig:ui}
\end{figure*}

%% file: references.bib
@misc{pawelczyk2021carla,
      title={CARLA: A Python Library to Benchmark Algorithmic Recourse and Counterfactual Explanation Algorithms},
      author={Martin Pawelczyk and Sascha Bielawski and Johannes van den Heuvel and Tobias Richter and Gjergji Kasneci},
      year={2021},
      eprint={2108.00783},
      archivePrefix={arXiv},
      primaryClass={cs.LG}
}

@article{Guo2024, doi = {10.21105/joss.06567}, url = {https://doi.org/10.21105/joss.06567}, year = {2024}, publisher = {The Open Journal}, volume = {9}, number = {103}, pages = {6567}, author = {Guo, Hangzhi and Xiong, Xinchang and Zhang, Wenbo and Yadav, Amulya}, title = {ReLax: Efficient and Scalable Recourse Explanation Benchmarking using JAX}, journal = {Journal of Open Source Software} 
}

@article{kleinberg2018human,
  title={Human decisions and machine predictions},
  author={Kleinberg, Jon and Lakkaraju, Himabindu and Leskovec, Jure and Ludwig, Jens and Mullainathan, Sendhil},
  journal={The quarterly journal of economics},
  volume={133},
  number={1},
  pages={237--293},
  year={2018},
  publisher={Oxford University Press}
}

@article{belz2025qra++,
  title={Qra++: Quantified reproducibility assessment for common types of results in natural language processing},
  author={Belz, Anya},
  journal={arXiv preprint arXiv:2505.17043},
  year={2025}
}

@article{voigt2017eu,
  title={The eu general data protection regulation (gdpr)},
  author={Voigt, Paul and Von dem Bussche, Axel},
  journal={A practical guide, 1st ed., Cham: Springer International Publishing},
  volume={10},
  number={3152676},
  pages={10--5555},
  year={2017},
  publisher={Springer}
}

@article{marzari2025probabilistically,
  title={Probabilistically robust counterfactual explanations under model changes},
  author={Marzari, Luca and Leofante, Francesco and Cicalese, Ferdinando and Farinelli, Alessandro},
  journal={Artificial Intelligence},
  pages={104459},
  year={2025},
  publisher={Elsevier}
}

@inproceedings{pawelczyk2020learning,
  title={Learning model-agnostic counterfactual explanations for tabular data},
  author={Pawelczyk, Martin and Broelemann, Klaus and Kasneci, Gjergji},
  booktitle={Proceedings of the web conference 2020},
  pages={3126--3132},
  year={2020}
}

@inproceedings{Wang2025,
author = {Wang, Haotian and Zou, Hao and Zhou, Xueguang and Wang, Shangwen and Yang, Wenjing and Cui, Peng},
title = {Learning Feasible Causal Algorithmic Recourse: A Prior Structural Knowledge Free Approach},
year = {2025},
isbn = {9798400712746},
publisher = {Association for Computing Machinery},
address = {New York, NY, USA},
url = {https://doi.org/10.1145/3696410.3714859},
doi = {10.1145/3696410.3714859},
booktitle = {Proceedings of the ACM on Web Conference 2025},
pages = {4507--4518},
numpages = {12},
location = {Sydney NSW, Australia},
series = {WWW '25}
}

@inproceedings{
pegios2025,
title={Counterfactual Explanations via Riemannian Latent Space Traversal},
author={Paraskevas Pegios and Aasa Feragen and Andreas Abildtrup Hansen and Georgios Arvanitidis},
booktitle={NeurIPS 2024 Workshop on Symmetry and Geometry in Neural Representations},
year={2025}
}

@misc{samoilescu2021,
author = {Samoilescu, Robert and Looveren, Arnaud and Klaise, Janis},
year = {2021},
month = {06},
pages = {},
title = {Model-agnostic and Scalable Counterfactual Explanations via Reinforcement Learning},
doi = {10.48550/arXiv.2106.02597}
}

@inproceedings{nguyen2022robust,
  title={Robust bayesian recourse},
  author={Nguyen, Tuan-Duy H and Bui, Ngoc and Nguyen, Duy and Yue, Man-Chung and Nguyen, Viet Anh},
  booktitle={Uncertainty in Artificial Intelligence},
  pages={1498--1508},
  year={2022},
  organization={PMLR}
}

@inproceedings{tolomei2017interpretable,
  title={Interpretable predictions of tree-based ensembles via actionable feature tweaking},
  author={Tolomei, Gabriele and Silvestri, Fabrizio and Haines, Andrew and Lalmas, Mounia},
  booktitle={Proceedings of the 23rd ACM SIGKDD international conference on knowledge discovery and data mining},
  pages={465--474},
  year={2017}
}

@article{laugel2017inverse,
  title={Inverse classification for comparison-based interpretability in machine learning},
  author={Laugel, Thibault and Lesot, Marie-Jeanne and Marsala, Christophe and Renard, Xavier and Detyniecki, Marcin},
  journal={arXiv preprint arXiv:1712.08443},
  year={2017}
}

@article{bui2025coverage,
  title={Coverage-validity-aware algorithmic recourse},
  author={Bui, Ngoc and Nguyen, Duy and Yue, Man-Chung and Nguyen, Viet Anh},
  journal={Operations Research},
  volume={73},
  number={6},
  pages={3294--3310},
  year={2025},
  publisher={INFORMS}
}

@inproceedings{leofante2024promoting,
  title={Promoting Counterfactual Robustness Through Diversity},
  author={Leofante, Francesco and Potyka, Nico},
  booktitle={Proceedings of the AAAI Conference on Artificial Intelligence},
  volume={38},
  pages={21322--21330},
  year={2024}
}

@inproceedings{poyiadzi2020face,
  title={FACE: feasible and actionable counterfactual explanations},
  author={Poyiadzi, Rafael and Sokol, Kacper and Santos-Rodriguez, Raul and De Bie, Tijl and Flach, Peter},
  booktitle={Proceedings of the AAAI/ACM Conference on AI, Ethics, and Society},
  pages={344--350},
  year={2020}
}

@article{mahajan2019preserving,
  title={Preserving causal constraints in counterfactual explanations for machine learning classifiers},
  author={Mahajan, Divyat and Tan, Chenhao and Sharma, Amit},
  journal={arXiv preprint arXiv:1912.03277},
  year={2019}
}

@article{adult_dataset,
  title={UCI machine learning repository},
  author={Dua, Dheeru and Graff, Casey and others},
  year={2017},
  publisher={Beijing China}
}

@misc{german_dataset,
  author       = {Hofmann, Hans},
  title        = {{Statlog (German Credit Data)}},
  year         = {1994},
  howpublished = {UCI Machine Learning Repository},
  note         = {{DOI}: https://doi.org/10.24432/C5NC77}
}

@article{tominaga2024,
  title={Reassessing evaluation functions in algorithmic recourse: An empirical study from a human-centered perspective},
  author={Tominaga, Tomu and Yamashita, Naomi and Kurashima, Takeshi},
  journal={arXiv preprint arXiv:2405.14264},
  year={2024}
}

@misc{canada_directive_adm_2025,
  author       = {{Government of Canada}},
  title        = {Directive on Automated Decision-Making},
  year         = {2025},
  url          = {https://www.tbs-sct.canada.ca/pol/doc-eng.aspx?id=32592},
  note         = {Last modified: 2025-06-24},
  urldate      = {2026-08-25}
}

@misc{corrected_german_dataset,
  title        = {{South German Credit}},
  author       = {Open Data LMU},
  year         = {2019},
  howpublished = {UCI Machine Learning Repository},
  note         = {{DOI}: https://doi.org/10.24432/C5X89F}
}

@article{Boston_housing_dataset,
  title = {Hedonic prices and the demand for clean air},
  author = {Harrison, David and Rubinfeld, Daniel L.},
  journal = {Journal of Environmental Economics and Management},
  volume = {5},
  number = {1},
  pages = {81--102},
  year = {1978},
  publisher = {Elsevier},
  doi = {10.1016/0095-0696(78)90006-2}
}

@misc{Credit_dataset,
    author = {Credit Fusion and Will Cukierski},
    title = {Give Me Some Credit},
    year = {2011},
    howpublished = {\url{https://kaggle.com/competitions/GiveMeSomeCredit}},
    note = {Kaggle}
}

@article{compas_dataset,
  author    = {Angwin, Julia and Larson, Jeff and Mattu, Surya and Kirchner, Lauren},
  title     = {Machine Bias: Risk Assessments in Criminal Sentencing},
  journal   = {ProPublica},
  year      = {2016},
  month     = {May},
  day       = {23},
  url       = {https://www.propublica.org/article/machine-bias-risk-assessments-in-criminal-sentencing}
}

@article{joshi2019towards,
  title={Towards realistic individual recourse and actionable explanations in black-box decision making systems},
  author={Joshi, Shalmali and Koyejo, Oluwasanmi and Vijitbenjaronk, Warut and Kim, Been and Ghosh, Joydeep},
  journal={arXiv preprint arXiv:1907.09615},
  year={2019}
}

@article{jiang2025robustx,
  title={Robustx: Robust counterfactual explanations made easy},
  author={Jiang, Junqi and Marzari, Luca and Purohit, Aaryan and Leofante, Francesco},
  journal={arXiv preprint arXiv:2502.13751},
  year={2025}
}

@article{dwivedi2023explainable,
  title={Explainable AI (XAI): Core ideas, techniques, and solutions},
  author={Dwivedi, Rudresh and Dave, Devam and Naik, Het and Singhal, Smiti and Omer, Rana and Patel, Pankesh and Qian, Bin and Wen, Zhenyu and Shah, Tejal and Morgan, Graham and others},
  journal={ACM computing surveys},
  volume={55},
  number={9},
  pages={1--33},
  year={2023},
  publisher={ACM New York, NY}
}

@article{Altmeyer2023,
  doi = {10.21105/jcon.00130},
  url = {https://doi.org/10.21105/jcon.00130},
  year = {2023},
  publisher = {The Open Journal},
  volume = {1},
  number = {1},
  pages = {130},
  author = {Patrick Altmeyer and Arie van Deursen and Cynthia C. S. Liem},
  title = {Explaining Black-Box Models through Counterfactuals},
  journal = {Proceedings of the JuliaCon Conferences}
}

@article{alibi2021,
  author  = {Janis Klaise and Arnaud Van Looveren and Giovanni Vacanti and Alexandru Coca},
  title   = {Alibi Explain: Algorithms for Explaining Machine Learning Models},
  journal = {Journal of Machine Learning Research},
  year    = {2021},
  volume  = {22},
  number  = {181},
  pages   = {1-7},
  url     = {http://jmlr.org/papers/v22/21-0017.html}
}

@article{nuredin2024impact,
  title={The impact of AI-based decision-making systems on justice and equality},
  author={Nuredin, Abdulmecit and Inan, Tevfik Can},
  journal={International Scientific Journal Vision},
  pages={9--33},
  year={2024}
}

@article{khosravi2024artificial,
  title={Artificial intelligence and decision-making in healthcare: a thematic analysis of a systematic review of reviews},
  author={Khosravi, Mohsen and Zare, Zahra and Mojtabaeian, Seyyed Morteza and Izadi, Reyhane},
  journal={Health services research and managerial epidemiology},
  volume={11},
  pages={23333928241234863},
  year={2024},
  publisher={SAGE Publications Sage CA: Los Angeles, CA}
}

@article{dote2025leveraging,
  title={Leveraging artificial intelligence for enhanced decision-making in finance: trends and future directions},
  author={Dote-Pardo, Jairo Stefano and Cordero-D{\'\i}az, Marling Carolina and Espinosa Jaramillo, Maria Teresa and Parra-Dom{\'\i}nguez, Javier},
  journal={Journal of Accounting Literature},
  year={2025},
  publisher={Emerald Publishing Limited}
}

@misc{jax2018github,
  author = {James Bradbury and Roy Frostig and Peter Hawkins and Matthew James Johnson and Yash Katariya and Chris Leary and Dougal Maclaurin and George Necula and Adam Paszke and Jake Vander{P}las and Skye Wanderman-{M}ilne and Qiao Zhang},
  title = {{JAX}: composable transformations of {P}ython+{N}um{P}y programs},
  url = {http://github.com/jax-ml/jax},
  version = {0.3.13},
  year = {2018},
}

@article{downs2020cruds,
  title={Cruds: Counterfactual recourse using disentangled subspaces},
  author={Downs, Michael and Chu, Jonathan L and Yacoby, Yaniv and Doshi-Velez, Finale and Pan, Weiwei},
  journal={ICML WHI},
  volume={2020},
  pages={1--23},
  year={2020}
}

@article{black2021consistent,
  title={Consistent counterfactuals for deep models},
  author={Black, Emily and Wang, Zifan and Fredrikson, Matt and Datta, Anupam},
  journal={arXiv preprint arXiv:2110.03109},
  year={2021}
}

@book{EU2024AIAct,
author = {European Data Protection Supervisor},
title = {AI Act Regulation (EU) 2024/1689 – Regulation (EU) 2024/1689 of the European Parliament and of the Council of 13 June 2024 laying down harmonised rules on artificial intelligence and amending Regulations (EC) No 300/2008, (EU) No 167/2013, (EU) No 168/2013, (EU) 2018/858, (EU) 2018/1139 and (EU) 2019/2144 and Directives 2014/90/EU, (EU) 2016/797 and (EU) 2020/1828 (Artificial Intelligence Act) (Text with EEA relevance)},
publisher = {Publications Office of the European Union},
year = {2025},
doi = {doi/10.2804/4225375}
}


%% file: zahra_references.bib
@misc{jiang_argumentative_2025,
	title = {Argumentative {Ensembling} for {Robust} {Recourse} under {Model} {Multiplicity}},
	url = {http://arxiv.org/abs/2506.20260},
	doi = {10.48550/arXiv.2506.20260},
	urldate = {2026-04-30},
	publisher = {arXiv},
	author = {Jiang, Junqi and Rago, Antonio and Leofante, Francesco and Toni, Francesca},
	month = jun,
	year = {2025},
	note = {arXiv:2506.20260 [cs]},
}

@article{hamman_robust_2024,
	title = {Robust {Algorithmic} {Recourse} {Under} {Model} {Multiplicity} {With} {Probabilistic} {Guarantees}},
	volume = {5},
	issn = {2641-8770},
	url = {https://ieeexplore.ieee.org/document/10530968/},
	doi = {10.1109/JSAIT.2024.3401407},
	urldate = {2026-04-30},
	journal = {IEEE Journal on Selected Areas in Information Theory},
	author = {Hamman, Faisal and Noorani, Erfaun and Mishra, Saumitra and Magazzeni, Daniele and Dutta, Sanghamitra},
	year = {2024},
	pages = {357--368},
}

@misc{yadav_low-cost_2022,
	title = {Low-{Cost} {Algorithmic} {Recourse} for {Users} {With} {Uncertain} {Cost} {Functions}},
	url = {http://arxiv.org/abs/2111.01235},
	doi = {10.48550/arXiv.2111.01235},
	urldate = {2026-04-30},
	publisher = {arXiv},
	author = {Yadav, Prateek and Hase, Peter and Bansal, Mohit},
	month = feb,
	year = {2022},
	note = {arXiv:2111.01235 [cs]},
}

@misc{kayastha_learning-augmented_2026,
	title = {Learning-{Augmented} {Robust} {Algorithmic} {Recourse}},
	url = {http://arxiv.org/abs/2410.01580},
	doi = {10.48550/arXiv.2410.01580},
	urldate = {2026-04-30},
	publisher = {arXiv},
	author = {Kayastha, Kshitij and Gkatzelis, Vasilis and Jabbari, Shahin},
	month = apr,
	year = {2026},
	note = {arXiv:2410.01580 [cs]},
}

@misc{antoran_getting_2021,
	title = {Getting a {CLUE}: {A} {Method} for {Explaining} {Uncertainty} {Estimates}},
	shorttitle = {Getting a {CLUE}},
	url = {http://arxiv.org/abs/2006.06848},
	doi = {10.48550/arXiv.2006.06848},
	urldate = {2026-04-30},
	publisher = {arXiv},
	author = {Antorán, Javier and Bhatt, Umang and Adel, Tameem and Weller, Adrian and Hernández-Lobato, José Miguel},
	month = mar,
	year = {2021},
	note = {arXiv:2006.06848 [stat]},
}

@inproceedings{altmeyer_endogenous_2023,
	address = {Raleigh, NC, USA},
	title = {Endogenous {Macrodynamics} in {Algorithmic} {Recourse}},
	copyright = {https://doi.org/10.15223/policy-029},
	isbn = {978-1-6654-6299-0},
	url = {https://ieeexplore.ieee.org/document/10136130/},
	doi = {10.1109/SaTML54575.2023.00036},
	language = {en},
	urldate = {2026-04-30},
	booktitle = {2023 {IEEE} {Conference} on {Secure} and {Trustworthy} {Machine} {Learning} ({SaTML})},
	publisher = {IEEE},
	author = {Altmeyer, Patrick and Angela, Giovan and Buszydlik, Aleksander and Dobiczek, Karol and Van Deursen, Arie and Liem, Cynthia C. S.},
	month = feb,
	year = {2023},
	pages = {418--431},
}

@article{semmelrock_reproducibility_2025,
	title = {Reproducibility in machine-learning-based research: {Overview}, barriers, and drivers},
	volume = {46},
	number = {2},
	journal = {AI Magazine},
	publisher = {Wiley Online Library},
	author = {Semmelrock, Harald and Ross-Hellauer, Tony and Kopeinik, Simone and Theiler, Dieter and Haberl, Armin and Thalmann, Stefan and Kowald, Dominik},
	year = {2025},
	pages = {e70002},
}

@book{sciences_reproducibility_2019,
	title = {Reproducibility and replicability in science},
	publisher = {National Academies Press},
	author = {Sciences, National Academies of and {Medicine} and {Policy} and Affairs, Global and Data, Board on Research and {Information} and Engineering, Division on and Sciences, Physical and Applied, Committee on and Statistics, Theoretical and {others}},
	year = {2019},
}

@article{rothermel_analyzing_1996,
	title = {Analyzing {Regression} {Test} {Selection} {Techniques}},
	volume = {22},
	issn = {0098-5589},
	url = {https://doi.org/10.1109/32.536955},
	doi = {10.1109/32.536955},
	number = {8},
	urldate = {2026-04-20},
	journal = {IEEE Trans. Softw. Eng.},
	author = {Rothermel, Gregg and Harrold, Mary Jean},
	month = aug,
	year = {1996},
	pages = {529--551},
}

@misc{desai_what_2025,
	title = {What is {Reproducibility} in {Artificial} {Intelligence} and {Machine} {Learning} {Research}?},
	url = {http://arxiv.org/abs/2407.10239},
	doi = {10.48550/arXiv.2407.10239},
	urldate = {2026-04-20},
	publisher = {arXiv},
	author = {Desai, Abhyuday and Abdelhamid, Mohamed and Padalkar, Nakul R.},
	month = mar,
	year = {2025},
	note = {arXiv:2407.10239 [cs]},
}

@inproceedings{henderson_deep_2018,
	address = {New Orleans, Louisiana, USA},
	series = {{AAAI}'18/{IAAI}'18/{EAAI}'18},
	title = {Deep reinforcement learning that matters},
	isbn = {978-1-57735-800-8},
	url = {https://dl.acm.org/doi/10.5555/3504035.3504427},
	urldate = {2026-04-19},
	booktitle = {Proceedings of the {Thirty}-{Second} {AAAI} {Conference} on {Artificial} {Intelligence} and {Thirtieth} {Innovative} {Applications} of {Artificial} {Intelligence} {Conference} and {Eighth} {AAAI} {Symposium} on {Educational} {Advances} in {Artificial} {Intelligence}},
	publisher = {AAAI Press},
	author = {Henderson, Peter and Islam, Riashat and Bachman, Philip and Pineau, Joelle and Precup, Doina and Meger, David},
	month = feb,
	year = {2018},
	pages = {3207--3214},
}

@article{gundersen_machine_2022,
	title = {Do machine learning platforms provide out-of-the-box reproducibility?},
	volume = {126},
	issn = {0167-739X},
	url = {https://www.sciencedirect.com/science/article/pii/S0167739X21002090},
	doi = {10.1016/j.future.2021.06.014},
	urldate = {2026-04-20},
	journal = {Future Generation Computer Systems},
	author = {Gundersen, Odd Erik and Shamsaliei, Saeid and Isdahl, Richard Juul},
	month = jan,
	year = {2022},
	pages = {34--47},
}

@article{pineau2021improving,
  title={Improving reproducibility in machine learning research (a report from the neurips 2019 reproducibility program)},
  author={Pineau, Joelle and Vincent-Lamarre, Philippe and Sinha, Koustuv and Larivi{\`e}re, Vincent and Beygelzimer, Alina and d'Alch{\'e}-Buc, Florence and Fox, Emily and Larochelle, Hugo},
  journal={Journal of machine learning research},
  volume={22},
  number={164},
  pages={1--20},
  year={2021}
}

@article{baker_1500_2016,
	title = {1,500 scientists lift the lid on reproducibility},
	volume = {533},
	issn = {1476-4687},
	doi = {10.1038/533452a},
	language = {eng},
	number = {7604},
	journal = {Nature},
	author = {Baker, Monya},
	month = may,
	year = {2016},
	pages = {452--454},
}

@article{gundersen_state_2018,
	title = {State of the {Art}: {Reproducibility} in {Artificial} {Intelligence}},
	volume = {32},
	copyright = {Copyright (c)},
	issn = {2374-3468},
	shorttitle = {State of the {Art}},
	url = {https://ojs.aaai.org/index.php/AAAI/article/view/11503},
	doi = {10.1609/aaai.v32i1.11503},
	language = {en},
	number = {1},
	urldate = {2026-04-20},
	journal = {Proceedings of the AAAI Conference on Artificial Intelligence},
	author = {Gundersen, Odd Erik and Kjensmo, Sigbjørn},
	month = apr,
	year = {2018},
}

@inproceedings{mothilal_explaining_2020,
	title = {Explaining machine learning classifiers through diverse counterfactual explanations},
	booktitle = {Proceedings of the 2020 {Conference} on {Fairness}, {Accountability}, and {Transparency}},
	author = {Mothilal, Ramaravind K and Sharma, Amit and Tan, Chenhao},
	year = {2020},
	pages = {607--617},
}

@inproceedings{jiang_provably_2024,
	title = {Provably {Robust} and {Plausible} {Counterfactual} {Explanations} for {Neural} {Networks} via {Robust} {Optimisation}},
	issn = {2640-3498},
	url = {https://proceedings.mlr.press/v222/jiang24a.html},
	language = {en},
	urldate = {2025-11-07},
	booktitle = {Proceedings of the 15th {Asian} {Conference} on {Machine} {Learning}},
	publisher = {PMLR},
	author = {Jiang, Junqi and Lan, Jianglin and Leofante, Francesco and Rago, Antonio and Toni, Francesca},
	month = feb,
	year = {2024},
	pages = {582--597},
}

@article{virgolin_robustness_2023,
	title = {On the robustness of sparse counterfactual explanations to adverse perturbations},
	volume = {316},
	issn = {0004-3702},
	url = {https://www.sciencedirect.com/science/article/pii/S0004370222001801},
	doi = {10.1016/j.artint.2022.103840},
	urldate = {2025-11-03},
	journal = {Artificial Intelligence},
	author = {Virgolin, Marco and Fracaros, Saverio},
	month = mar,
	year = {2023},
	pages = {103840},
}

@inproceedings{wang_flexible_2023,
	address = {New York, NY, USA},
	series = {{CIKM} '23},
	title = {Flexible and {Robust} {Counterfactual} {Explanations} with {Minimal} {Satisfiable} {Perturbations}},
	isbn = {979-8-4007-0124-5},
	url = {https://dl.acm.org/doi/10.1145/3583780.3614885},
	doi = {10.1145/3583780.3614885},
	urldate = {2025-11-03},
	booktitle = {Proceedings of the 32nd {ACM} {International} {Conference} on {Information} and {Knowledge} {Management}},
	publisher = {Association for Computing Machinery},
	author = {Wang, Yongjie and Qian, Hangwei and Liu, Yongjie and Guo, Wei and Miao, Chunyan},
	month = oct,
	year = {2023},
	pages = {2596--2605},
}

@misc{pawelczyk_probabilistically_2023,
	title = {Probabilistically {Robust} {Recourse}: {Navigating} the {Trade}-offs between {Costs} and {Robustness} in {Algorithmic} {Recourse}},
	shorttitle = {Probabilistically {Robust} {Recourse}},
	url = {http://arxiv.org/abs/2203.06768},
	doi = {10.48550/arXiv.2203.06768},
	urldate = {2025-11-03},
	publisher = {arXiv},
	author = {Pawelczyk, Martin and Datta, Teresa and van-den-Heuvel, Johannes and Kasneci, Gjergji and Lakkaraju, Himabindu},
	month = oct,
	year = {2023},
	note = {arXiv:2203.06768 [cs]},
}

@misc{upadhyay_towards_2021,
	title = {Towards {Robust} and {Reliable} {Algorithmic} {Recourse}},
	url = {http://arxiv.org/abs/2102.13620},
	doi = {10.48550/arXiv.2102.13620},
	urldate = {2025-11-03},
	publisher = {arXiv},
	author = {Upadhyay, Sohini and Joshi, Shalmali and Lakkaraju, Himabindu},
	month = jul,
	year = {2021},
	note = {arXiv:2102.13620 [cs]},
}

@article{wachter_counterfactual_2017,
	title = {Counterfactual {Explanations} {Without} {Opening} the {Black} {Box}: {Automated} {Decisions} and the {GDPR}},
	issn = {1556-5068},
	shorttitle = {Counterfactual {Explanations} {Without} {Opening} the {Black} {Box}},
	url = {https://www.ssrn.com/abstract=3063289},
	doi = {10.2139/ssrn.3063289},
	language = {en},
	urldate = {2025-08-11},
	journal = {SSRN Electronic Journal},
	author = {Wachter, Sandra and Mittelstadt, Brent and Russell, Chris},
	year = {2017},
}

@inproceedings{ustun_actionable_2019,
	title = {Actionable recourse in linear classification},
	booktitle = {Proceedings of the conference on fairness, accountability, and transparency},
	author = {Ustun, Berk and Spangher, Alexander and Liu, Yang},
	year = {2019},
	pages = {10--19},
}

@article{karimi_survey_2022,
	title = {A {Survey} of {Algorithmic} {Recourse}: {Contrastive} {Explanations} and {Consequential} {Recommendations}},
	volume = {55},
	issn = {0360-0300},
	shorttitle = {A {Survey} of {Algorithmic} {Recourse}},
	url = {https://dl.acm.org/doi/10.1145/3527848},
	doi = {10.1145/3527848},
	number = {5},
	urldate = {2025-04-08},
	journal = {ACM Comput. Surv.},
	author = {Karimi, Amir-Hossein and Barthe, Gilles and Schölkopf, Bernhard and Valera, Isabel},
	month = dec,
	year = {2022},
	pages = {95:1--95:29},
}
